\pdfoutput=1
\documentclass[letterpaper]{article} 
\usepackage{aaai23}  
\usepackage{times}  
\usepackage{helvet}  
\usepackage{courier}  
\usepackage[hyphens]{url}  
\usepackage{graphicx} 
\usepackage{natbib}  
\usepackage{caption} 
\usepackage{algorithm}
\usepackage{algorithmic}

\usepackage{epsfig}
\usepackage{amssymb}
\usepackage{makecell}
\usepackage{graphicx}
\usepackage{amsmath}
\usepackage{amsthm}
\usepackage{mathrsfs}
\usepackage{amssymb}
\usepackage{booktabs}
\usepackage{algorithm}
\usepackage{algorithmic}
\usepackage{bm} 
\usepackage{overpic}                                                            
\usepackage{subfigure}
\usepackage{color}
\usepackage{threeparttable}
\usepackage{caption}

\usepackage{multirow}
\newtheorem{theorem}{Theorem}

\newtheorem{definition}{Definition}

\usepackage{newfloat}
\usepackage{listings}
\DeclareCaptionStyle{ruled}{labelfont=normalfont,labelsep=colon,strut=off} 
\floatstyle{ruled}
\newfloat{listing}{tb}{lst}{}
\floatname{listing}{Listing}
\title{Reducing ANN-SNN Conversion Error through Residual Membrane Potential}
\author {
    Zecheng Hao\textsuperscript{\rm 1},
    Tong Bu\textsuperscript{\rm 1,\rm 2},
    Jianhao Ding\textsuperscript{\rm 1},
    Tiejun Huang\textsuperscript{\rm 1,\rm 2},
    Zhaofei Yu\textsuperscript{\rm 1,\rm 2}\thanks{Corresponding author}
}
\affiliations {
    \textsuperscript{\rm 1} School of Computer Science, Peking University\\
    \textsuperscript{\rm 2} Institute for Artificial Intelligence, Peking University\\
    zechenghao@pku.edu.cn,
    putong30@pku.edu.cn, 
    djh01998@stu.pku.edu.cn, 
    \{tjhuang,yuzf12\}@pku.edu.cn
}

\begin{document}

\maketitle

\begin{abstract}
Spiking Neural Networks (SNNs) have received extensive academic attention due to the unique properties of low power consumption and high-speed computing on neuromorphic chips. Among various training methods of SNNs, ANN-SNN conversion has shown the equivalent level of performance as ANNs on large-scale datasets. However, unevenness error, which refers to the deviation caused by different temporal sequences of spike arrival on activation layers, has not been effectively resolved and seriously suffers the performance of SNNs under the condition of short time-steps. In this paper, we make a detailed analysis of unevenness error and divide it into four categories. We point out that the case of the ANN output being zero while the SNN output being larger than zero accounts for the largest percentage. Based on this, we theoretically prove the sufficient and necessary conditions of this case and propose an optimization strategy based on residual membrane potential to reduce unevenness error. The experimental results show that the proposed method achieves state-of-the-art performance on CIFAR-10, CIFAR-100, and ImageNet datasets. For example, we reach top-1 accuracy of 64.32\% on ImageNet with 10-steps. To the best of our knowledge, this is the first time ANN-SNN conversion can simultaneously achieve high accuracy and ultra-low-latency on the complex dataset. Code is available at https://github.com/hzc1208/ANN2SNN\_SRP.

\end{abstract}

\section{Introduction}
Spiking Neural Networks (SNNs), known as the third generation of artificial neural networks~\cite{Wolfgang1997}, have shown distinctive properties and remarkable advantages of temporal information processing capability and high biological plausibility~\cite{roy2019towards}. As each neuron delivers binary spikes only when the membrane potential reaches the threshold, the calculation in SNNs has the property of high-sparsity and multiplication-free, which brings unique advantages of low power consumption on neuromorphic chips~\cite{schemmel2010wafer,furber2012overview,merolla2014million,davies2018loihi,pei2019towards,debole2019truenorth}. 
Despite these, it remains challenging to train high-performance and low-latency SNNs.

Generally, two main approaches are proposed to train deep SNNs: (1) backpropagation with surrogate gradient~\cite{neftci2019surrogate,kim2020unifying}, (2) ANN-SNN conversion~\cite{cao2015spiking,rueckauer2016theory}. Unlike the surrogate gradient-based learning method that requires more GPU computing than ANN training, ANN-SNN conversion directly trains a source ANN and then converts it to an SNN by replacing the ReLU activation with IF neurons, which is regarded as the most effective way to train deep SNNs. 

Although ANN-SNN conversion can achieve comparable performance as ANNs on large-scale datasets, it often requires large time-steps to match the firing rates of SNNs to the activation value of ANNs. There is still a performance gap between ANNs and SNNs under the condition of low latency. Many researchers have made great efforts to eliminate these errors to achieve high-performance converted SNNs with low latency~\cite{han2020rmp,ding2021optimal,ho2021tcl,deng2020optimal,li2021free,li2022quantization}. However, unevenness error~\cite{bu2022optimal}, which refers to the deviation caused by different temporal sequences of spike arrival on activation layers, has not been effectively resolved and seriously suffers the performance of converted SNNs. 

In this paper, by analyzing and deducing the distribution of unevenness error, we propose an optimization strategy based on residual membrane potential (SRP), which can decrease unevenness error and improve network performance effectively under low latency. Our main contributions are summarized as follows:
\begin{itemize}
    \item[1] We divide unevenness error into four cases according to the outputs of ANNs and SNNs. We systematically analyze the distribution of unevenness error under these four cases and point out that the case of the ANN output being zero while the SNN output being larger than zero accounts for the largest percentage.
    \item[2] We theoretically establish the mathematical relationship between residual membrane potential and the specific case of unevenness error, and propose an optimization strategy based on residual membrane potential to reduce unevenness error.
    \item[3] We demonstrate the effectiveness of the proposed method on CIFAR-10/100, and ImageNet. Our method outperforms previous state-of-the-art and shows remarkable advantages on all tested datasets and network structures.
     \item[4] We show that our method is compatible with other ANN-SNN conversion methods and remarkably improves the performance when the time-steps are small.
\end{itemize}

\section{Related Work}
Due to the non-differentiable property of the spike firing mechanism, the training of SNNs is not as easy as ANNs. 
In recent years, two mainstream training methods have been proposed and widely used in obtaining deep SNNs.\\
\textbf{Supervised learning of SNNs}. To overcome the non-differentiable problem, smoothing and surrogate gradient methods \cite{huh2017gradient,jin2018hybrid,wu2018STBP,shrestha2018slayer,zenke2018superspike,bellec2018long,neftci2019surrogate} have been proposed. These models maintain the network architecture unchanged in forward propagation and replace the Heaviside function of membrane potential with a differentiable function in back-propagation. 
On this basis, many researchers began to improve the performance of SNNs by analyzing the surrogate gradients~\cite{Zenke2020.06.29.176925,li2021differentiable}, designing appropriate network structures~\cite{zheng2021going,fang2020incorporating} and loss functions~\cite{deng2022temporal,guo2022recdis}, and using trainable membrane time constant~\cite{fang2020incorporating}.
Event-driven learning is another type of learning method, which makes better use of the temporal information in SNNs based on the differentiation of spike firing time. SpikeProp~\cite{bohte2002error} was the first event-driven supervised learning algorithm for SNNs, which aimed to use linear smoothing approximation to overcome the non-differentiable problem of the relative membrane potential in spike firing time.  Kheradpisheh et al.~\shortcite{kheradpisheh2020temporal} set the gradient as $-1$ to train shallow networks. To improve temporal learning precision, Zhang et al.~\shortcite{zhang2020temporal} attempted to retain inter-neuron and intra-neuron dependencies. By calculating the 
presynaptic and postsynaptic spike firing time, Mostafa et al.~\shortcite{mostafa2017supervised} and Zhou et al.~\shortcite{zhou2021temporal}  constructed new continuous activation function, which directly avoids non-differentiable problem. Kim et al.~\shortcite{kim2020unifying} combined rate-coding and temporal-coding together in the back-propagation of SNNs. Despite these, the activation-based methods consume a lot of memory and computation, while the event-driven methods are limited to shallow networks.\\
\textbf{ANN-SNN Conversion}. ANN-SNN Conversion aims to map the parameters of the pretrained ANNs to SNNs and reduce the performance loss as much as possible.  Cao et al.~\shortcite{cao2015spiking} firstly trained ANNs with the ReLU activation function and then replaced the activation layers with spiking neurons. As the firing rate is not precise enough to match the output of respective activation layers in ANNs, researchers have proposed many optimization methods to reduce conversion error. Diehl et al.~\shortcite{diehl2015fast} attempted to narrow the gap between ANNs and SNNs by scaling and normalizing weights. Rueckauer et al.~\shortcite{Bodo2017Conversion} proposed the ``reset-by-subtraction" mechanism to retain the temporal information, which was also adopted in the following work~\cite{han2020rmp,deng2020optimal,bu2022optimal}. Besides,
many works focused on improving the performance by adopting various mechanism to dynamically adjust the threshold \cite{Bodo2017Conversion,sengupta2019going,han2020rmp,ding2021optimal,stockl2021optimized,ho2021tcl,wu2021tandem}. Kim et al.~\shortcite{kim2020spiking} pointed out that lossless conversion can be realized when the time-step is sufficiently long. Therefore, recent works focused on improving performance under the condition of short time-steps by optimizing the weights, biases~\cite{deng2020optimal,li2021free} and initial membrane potential~\cite{Bu2022OPI}, utilizing burst spikes~\cite{li2022efficient} and memory function~\cite{wang2022signed}, and designing quantization clip-floor-shift activation function for ANNs~\cite{bu2022optimal}. 
Although these methods can achieve high performance with few ($4-8$) time-steps on the CIFAR-10/100 datasets, they usually take large number of time-steps (usually $>16$) to achieve comparable performance as ANNs on complex dataset like ImageNet, due to the effects of unevenness error. This paper aims to reduce unevenness error and obtain high-accuracy and ultra-low-latency SNNs on complex dataset.

\section{Preliminary}
\subsection{Neuron Model}
\textbf{ANN neurons.} For ANNs, the output $\boldsymbol{a}^l$ of neurons in layer $l$ is realized by
a linear weighting and a nonlinear mapping:
\begin{align}
    \boldsymbol{a}^l=f(\boldsymbol{W}^{l}\boldsymbol{a}^{l-1}).
    \label{ann}
\end{align}
where $\boldsymbol{W}^{l}$ refers to the weights between layer $l$ and layer $l-1$. $f(\cdot)$ is the nonlinear activation function, which is often set as ReLU function.\\
\textbf{SNN neurons.} For SNNs, we consider the commonly used Integrate-and-Fire (IF) model~\cite{diehl2015fast,sengupta2019going,han2020rmp,deng2020optimal,bu2022optimal}, the dynamics of which can be described by:
\begin{align}
	\boldsymbol{v}^{l}(t) &= \boldsymbol{v}^{l}(t-1) + \boldsymbol{W}^{l}{\theta}^{l-1} \boldsymbol{s}^{l-1}(t)-{\theta}^{l} \boldsymbol{s}^{l}(t).  \label{equ02}
\end{align}
where $\boldsymbol{v}^{l}(t)$ denotes the membrane potential of neurons in layer $l$ at time-step $t$, $\boldsymbol{W}^{l}$ refers to the weights between layer $l$ and layer $l-1$, and ${\theta}^l$ is the firing threshold of neurons in layer $l$. $\boldsymbol{s}^{l}(t)$ denotes the binary output spikes of neurons in layer $l$ at time-step $t$, which is defined as:
\begin{align}
    \boldsymbol{s}^{l}(t) &= H(\boldsymbol{u}^{l}(t) - {\theta}^{l}). \label{equ03}
\end{align}
where $\boldsymbol{u}^{l}(t)=\boldsymbol{v}^{l}(t-1) + \boldsymbol{W}^{l} {\theta}^{l-1} \boldsymbol{s}^{l-1}(t)$ denotes the membrane potential of neurons before the trigger of a spike at time-step $t$, $H(\cdot)$ is the Heaviside step function. The output spike $\boldsymbol{s}^{l}(t)$
equals 1 if the membrane potential $\boldsymbol{u}^{l}(t)$ is larger than the threshold ${\theta}^l$ and otherwise 0. Note that in Eq.~\eqref{equ02}, we use the “reset-by-subtraction” mechanism~\cite{Bodo2017Conversion} to reduce information loss, which means the membrane potential $\boldsymbol{v}^l(t)$ is subtracted by the threshold value ${\theta}^l$ if the neuron fires.

\subsection{ANN-SNN Conversion}
\textbf{Relate ANN to SNN.} 
The idea of ANN-SNN conversion is to relate the ReLU activation of analog neurons in ANNs to  the firing rate (or postsynaptic potential) of spiking neurons in SNNs. Specifically, by summing Eq.~\eqref{equ02} from time-step 1 to $T$ and dividing $T$ on both sides, we have:
\begin{figure} [t]\centering    
\subfigure[] {
 \label{fig0101}     
\includegraphics[width=0.75\columnwidth,trim=300 150 300 180,clip]{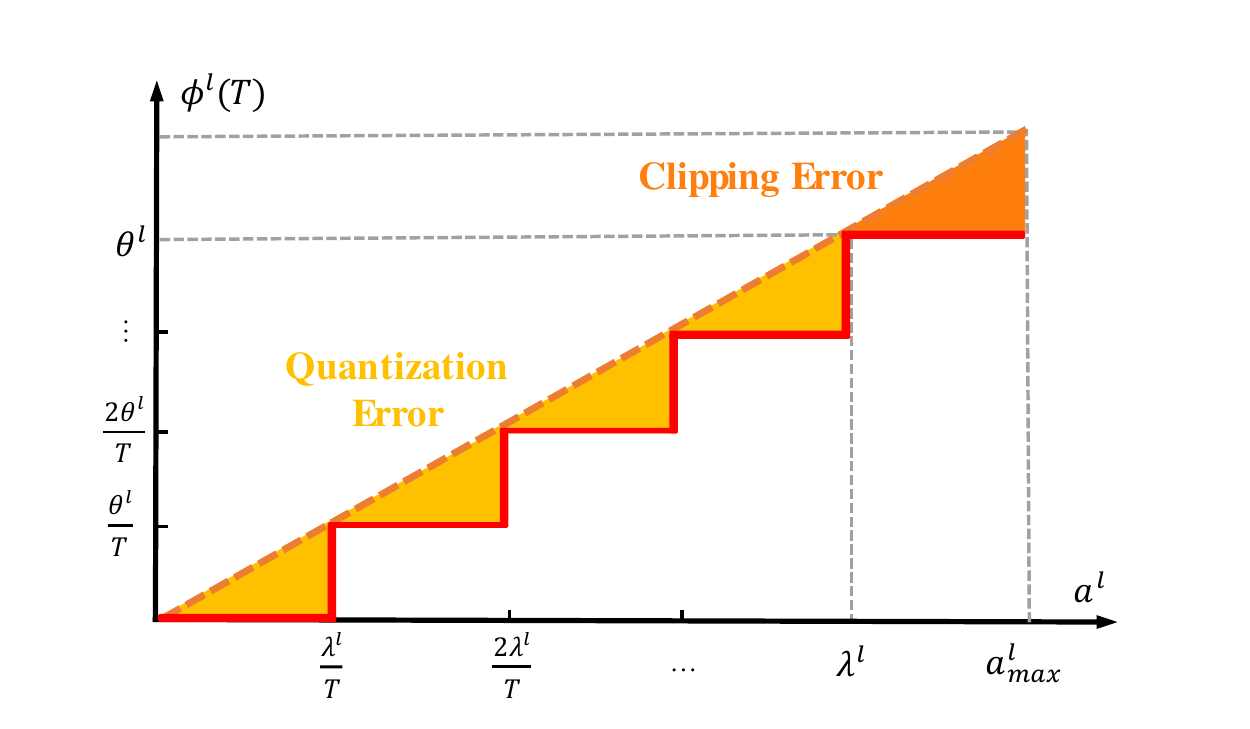}
}     
\subfigure[] { 
\label{fig0102}     
\includegraphics[width=0.30\columnwidth,trim=240 60 240 60,clip]{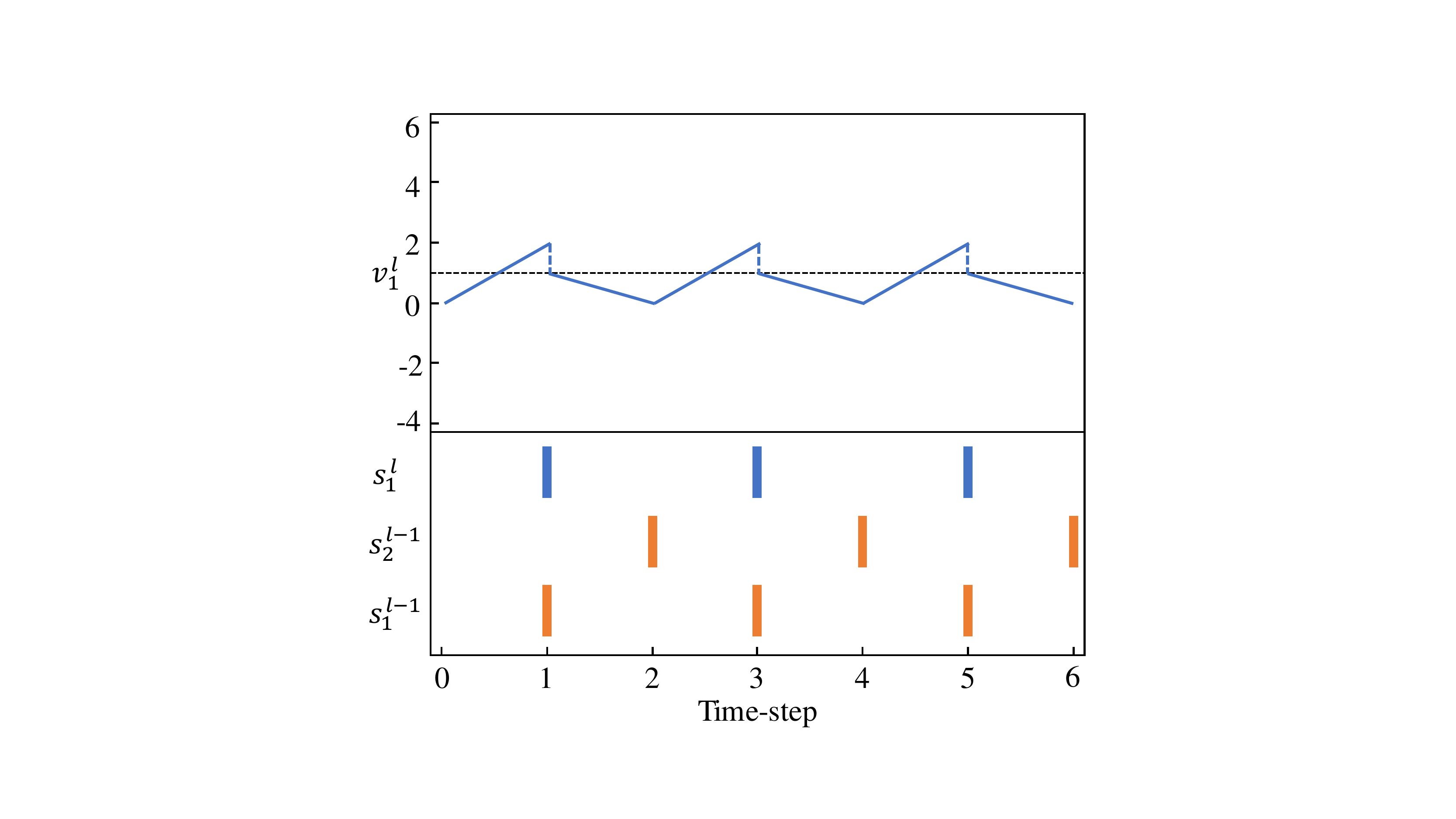}
}    
\subfigure[] { 
\label{fig0103}     
\includegraphics[width=0.30\columnwidth,trim=240 60 240 60,clip]{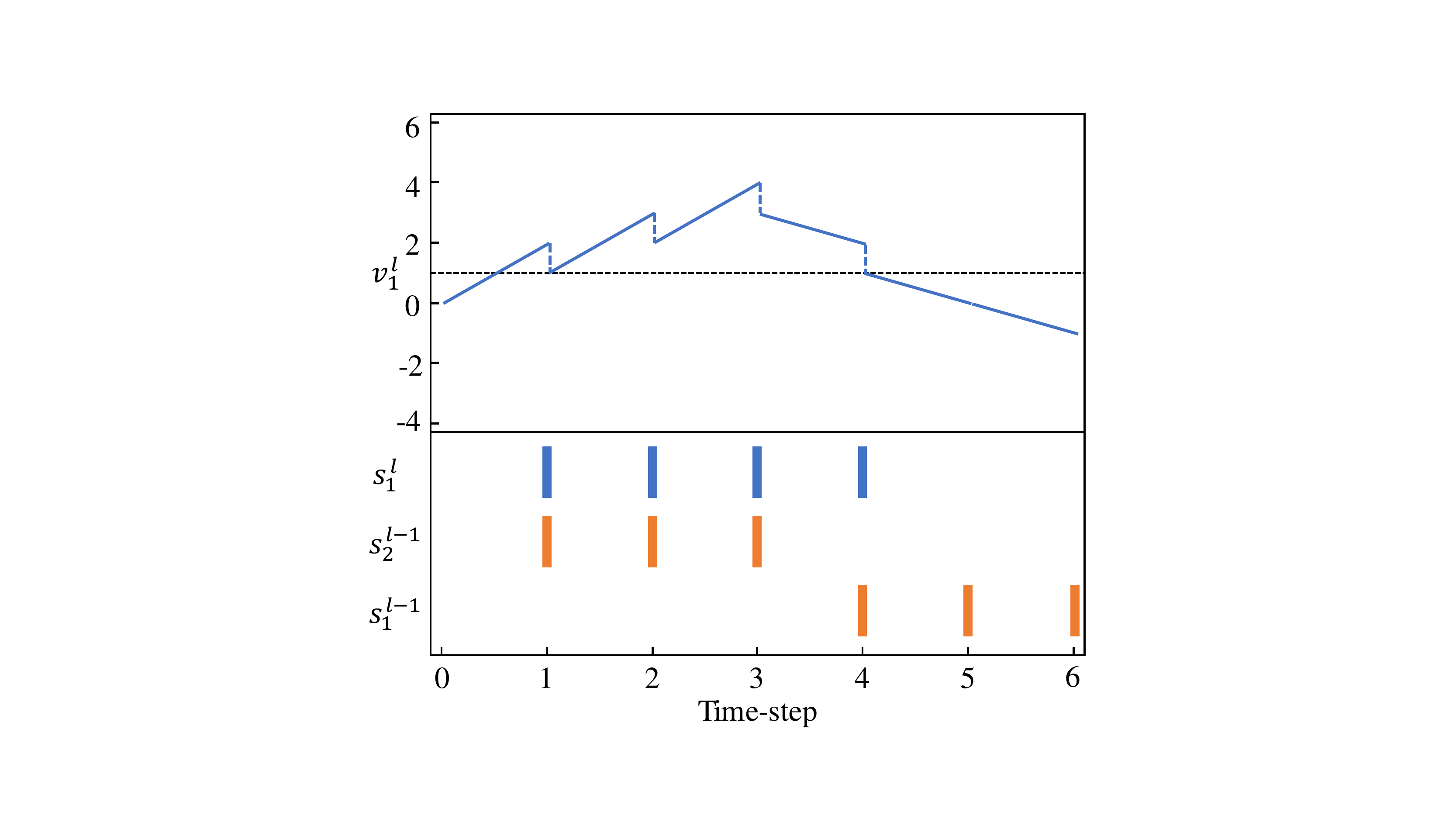} 
}   
\subfigure[] { 
\label{fig0104} 
\includegraphics[width=0.30\columnwidth,trim=240 60 240 60,clip]{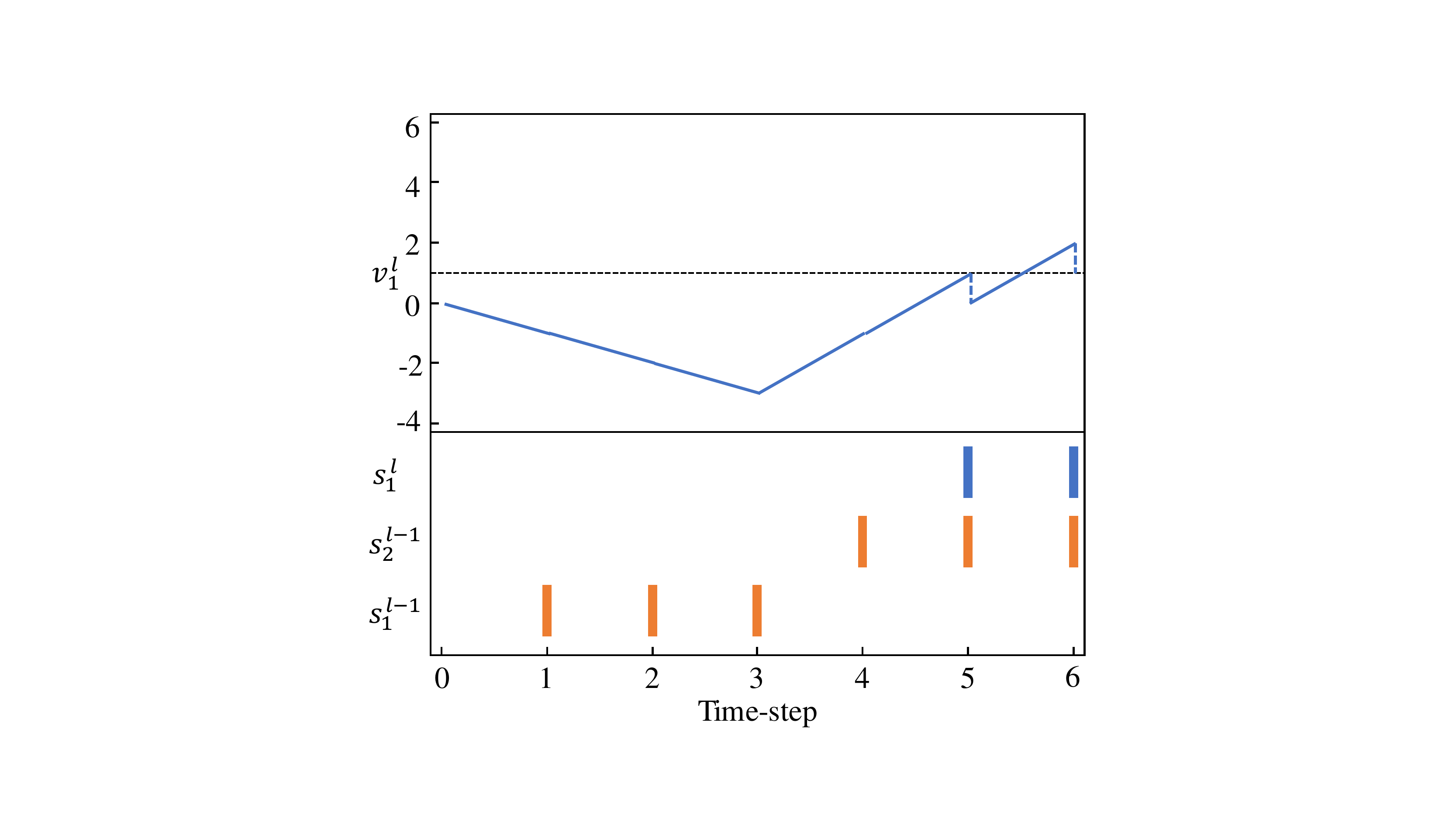}
}   
\caption{ANN-SNN conversion error includes: (a) Clipping error and Quantization error, (b)-(d) Unevenness error.}     
\label{fig01}     
\end{figure}
\begin{align}
    \frac{\boldsymbol{v}^{l}(T) - \boldsymbol{v}^{l}(0)}{T} &=  \frac{\sum\limits_{t=1}^T\boldsymbol{W}^{l}{\theta}^{l-1}\boldsymbol{s}^{l-1}(t)}{T} - \frac{\sum\limits_{t=1}^T{\theta}^{l} \boldsymbol{s}^{l}(t)}{T}. \label{equ06}
\end{align}
By  using $\boldsymbol{\phi}^l(T) = \frac{\sum\limits_{t=1}^T{\theta}^{l} \boldsymbol{s}^{l}(t)}{T}$ 
to denote the average postsynaptic potential, 
we will find the linear relationship between $\boldsymbol{\phi}^l(T)$ and $\boldsymbol{\phi}^{l-1}(T)$:
\begin{align}
      \boldsymbol{\phi}^l(T) &= \boldsymbol{W}^{l}\boldsymbol{\phi}^{l-1}(T) - \frac{\boldsymbol{v}^{l}(T) - \boldsymbol{v}^{l}(0)}{T}. \label{equ07}  
\end{align}
From Eqs.~\eqref{ann} and~\eqref{equ07}, we note that if we can map the activation value $\boldsymbol{a}^l$ of analog neurons in ANNs to $\boldsymbol{\phi}^l(T)$ of IF neurons in SNNs, then we will be able to train a source ANN with back-propagation (BP) and convert it to an SNN by replacing the ReLU activation with IF neurons, which is the core idea of ANN-SNN conversion. As Eqs.~\eqref{ann} and~\eqref{equ07} are not exactly equal, there exists conversion error in general.\\

\noindent
\textbf{ANN-SNN conversion error.}
There are three main errors in ANN-SNN conversion~\cite{bu2022optimal}, including clipping error, quantization error, and unevenness error, which cause the performance gap between ANNs and SNNs. Specifically,
\begin{itemize}
    \item[$\bullet$] \textbf{Clipping error} denotes the error caused by different value ranges of ANNs and SNNs. For ANNs, the output $\boldsymbol{a}^l$  belongs to a real number interval $[0,a_{max}^{l}]$ with $a_{max}^{l}$ denoting the maximum value of  $\boldsymbol{a}^l$. For SNNs,  the output $\boldsymbol{\phi}^l(T)$ belongs to a finite discrete set $S_T = \{ \frac{{\theta}^l i}{T} | i\in [0,T] \wedge  i\in \mathbb{N} \}$ due to $\boldsymbol{\phi}^l(T) = \sum\limits_{t=1}^T{\theta}^{l} \boldsymbol{s}^{l}(t) /T$. As illustrated in Fig.~\ref{fig01}(a), if we set $\lambda^l$ as an actual threshold in ANNs to map the maximum value ${\theta}^l$ in $S_T$ of SNNs, $\boldsymbol{a}^l$ can be mapped to $\boldsymbol{\phi}^l(T)$ by $\bm{\phi}^l(T)=\text{clip} \left(   \frac{ \theta^l}{T} \left \lfloor \frac{\bm{a}^{l} T}{\lambda^l}     \right \rfloor, 0, \theta^l  \right)$ with $\lfloor \cdot \rfloor$ denoting the floor function.
    Then the output $\boldsymbol{a}^l\in [\lambda^l, {a}_{max}^l]$ of the ANNs will be mapped to the same value ${\theta}^l$, which will cause the so-called clipping error.
    \item[$\bullet$] \textbf{Quantization error} is generated when mapping the continuous value of $\boldsymbol{a}^l$ in ANNs to the discrete value of $\boldsymbol{\phi}^l(T)$ in SNNs. As shown in Fig.~\ref{fig01}(a), when $\boldsymbol{a}^l\in[\frac{k\lambda^l}{T},\frac{(k+1)\lambda^l}{T})(k=0,1,...,T-1)$, the corresponding value of $\boldsymbol{\phi}^l(T)$ will always be $\frac{k{\theta}^l}{T}$.
    \item[$\bullet$] \textbf{Unevenness error} refers to the deviation between $\boldsymbol{a}^l$ and $\boldsymbol{\phi}^l(T)$ due to different temporal sequences of spike arrival on activation layers. Ideally, we expect the timing of receive spikes from the previous layer to be even. However, in reality, the spiking timing will be uneven when the spikes deliver to the deep layer, which will cause more spikes or fewer spikes than expected. An example is illustrated in Fig.~\ref{fig01}(b)-(d), in which two presynaptic neurons in layer $l-1$ are connected to a postsynaptic neuron in layer $l$. We assume $\boldsymbol{W}^l=\left[2\ -1\right], \boldsymbol{a}^{l-1} = \boldsymbol{\phi}^{l-1}(T) = \left[0.5\ 0.5\right]^T, {\theta}^l = \lambda^l = 1, T=6$, then we can get that the output $\boldsymbol{a}^{l}$ in ANN is $\boldsymbol{a}^{l}=\text{ReLU}(\boldsymbol{W}^l\boldsymbol{a}^{l-1})=0.5$.
    Under the ideal situation (Fig.~\ref{fig01}(b)), $\boldsymbol{\phi}^l(T) = \sum\limits_{t=1}^T{\theta}^{l} \boldsymbol{s}^{l}(t)/{T}=\frac{3}{6}=0.5$, which is exactly the same as ANN output. However, when the input spike timing changes
    (Fig.~\ref{fig01}(c)-(d)), $\phi^l(T)$ may become larger or smaller than $\boldsymbol{a}^l$ in fact.
\end{itemize}
\textbf{Eliminating conversion error.} The clipping and quantization errors can be eliminated by modifying the activation function of source ANNs. Specifically, the trainable clipping layer technique has been used to directly map the trainable upper bound of ANNs to the threshold of SNNs, so that the clipping error is zero~\cite{ho2021tcl}. Besides, the quantization clip-floor activation function was proposed to replace the ReLU activation function in source ANNs~\cite{bu2022optimal}, which can better approximate the activation function of SNNs and eliminate quantization error. Despite this, it is still challenging to reduce unevenness error, which seriously hampers the performance of SNNs under the condition of low time latency. Therefore, this paper aims to address unevenness error and explore high-performance
ANN-SNN conversion with ultra-low latency.

\section{Methods}
In this section, we make a detailed analysis of unevenness error and divide it into four situations, then we count out the distribution of unevenness error and point out that the first case plays a more important role among the four kinds of errors.
Based on this, we show the residual membrane potential is essential to ANN-SNN conversion and propose the optimization strategy to reduce unevenness error.

\subsection{Dividing Unevenness Error into Four Situations}
Here we divide unevenness error into different cases to analyze. We suppose that the ANN and SNN receive the same input to layer $l$, that is, $\boldsymbol{a}^{l-1}=\boldsymbol{\phi}^{l-1}(T)$, and then analyze the error between the ANN and SNN in layer $l$.  For simplicity, we use $\boldsymbol{y}^{l-1} = \boldsymbol{W}^{l} \boldsymbol{a}^{l-1}= \boldsymbol{W}^{l}\boldsymbol{\phi}^{l-1}(T)$
to substitute the weighted input to layer $l$ for both ANN and SNN. The conversion error is defined as the output of converted SNN subtracting the output of source ANN, which can be simplified according to Eqs.~\eqref{ann} and ~\eqref{equ07}:
\begin{align}
    \textbf{Error}^l&=\boldsymbol{\phi}^{l}(T)-\boldsymbol{a}^{l} \\
    &=\boldsymbol{y}^{l-1} - \frac{\boldsymbol{v}^{l}(T) - \boldsymbol{v}^{l}(0)}{T}-f(\boldsymbol{y}^{l-1}). \nonumber
\end{align}
In order to facilitate the analysis of unevenness error, we consider using the quantization clip-floor-shift (QCFS) function proposed in~\cite{bu2022optimal} to replace the ReLU activation function in source ANNs, that is:
\begin{align}
    f(\boldsymbol{y}^{l-1}) &= \lambda^l {\rm clip} \left(\frac{1}{L}\left\lfloor\frac{\boldsymbol{y}^{l-1} L}{\lambda^l}+\frac{1}{2}\right\rfloor,0,1 \right). \label{equ08} 
\end{align}
where $L$ is the quantization step of ANNs, $\lambda^l$ represents the trainable threshold in $l$-th layer of ANNs and is generally mapped to the threshold of SNN, that is, ${\theta}^l=\lambda^l$. The QCFS activation function better approximates the activation function of SNNs, and therefore could eliminate clipping and quantization errors (see Appendix for the detailed proof). Thus, the error ($\textbf{Error}^l$) here is caused by unevenness error.

\begin{figure} [t]\centering    
\includegraphics[width=0.75\columnwidth,trim=300 150 300 165,clip]{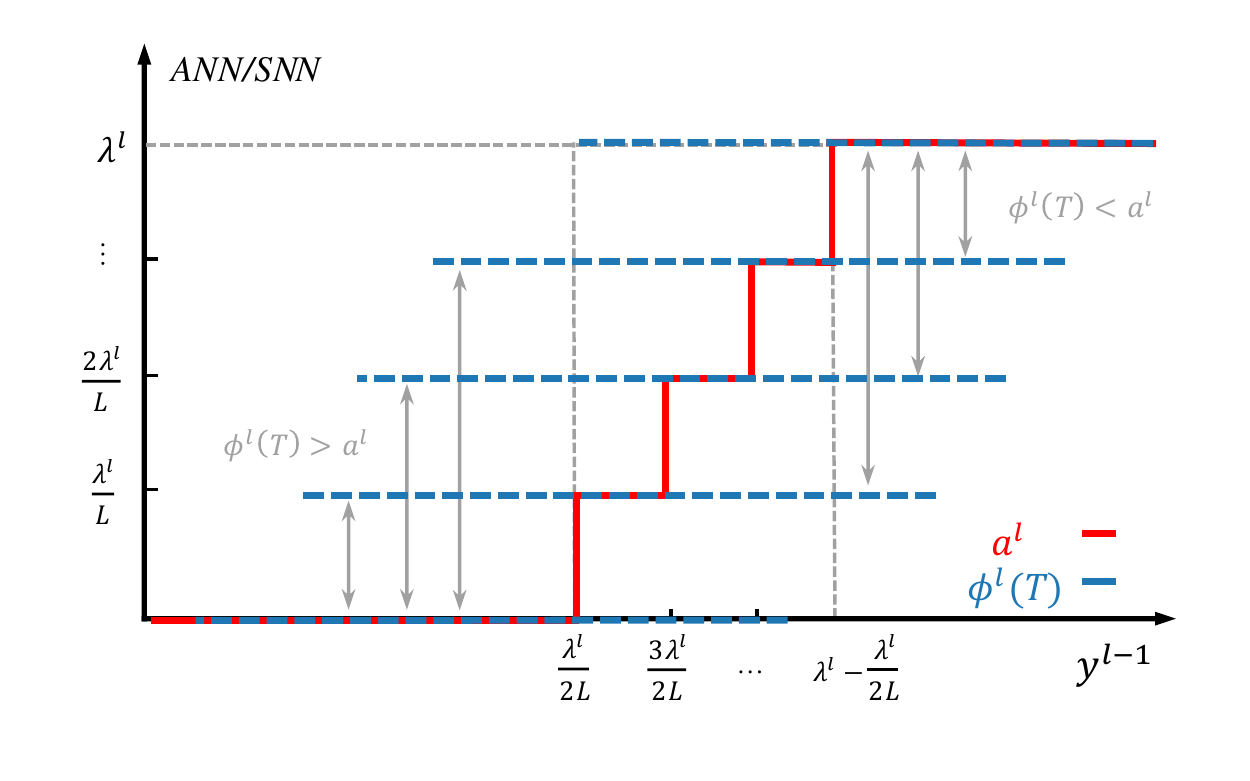}
\caption{Four cases of unevenness error.}
\label{fig02}     
\end{figure}

For further analysis, we compare the output $\boldsymbol{a}^{l}$ of ANN and $\boldsymbol{\phi}^{l}(T)$ of SNN in layer $l$ when receiving the same input $\boldsymbol{y}^{l-1}$.
As shown in  Fig.~\ref{fig02}, 
according to the  weighted input $\boldsymbol{y}^{l-1}$ to layer $l$ and the sign (positive and negative) of $\textbf{Error}^l$, unevenness error can be divided into the following four categories:
\begin{itemize}
    \item[$\bullet$] \textbf{Case 1}: ${a}^l=0, \phi^l(T)>{a}^l$, which means that the input ${y}^{l-1}< \lambda^l/2L$, the output of ANN equals 0 but the output of SNN is larger than 0 (fires more spikes as expected).
     \item[$\bullet$] \textbf{Case 2}: $0<{a}^l<\lambda^l,\phi^l(T)>{a}^l$, which means that the input $\lambda^l/2L \leqslant {y}^{l-1}<\lambda^l-\lambda^l/2L$, the output of SNN is larger than that of ANN (fires more spikes as expected).
     \item[$\bullet$] \textbf{Case 3}: $0<{a}^l<\lambda^l,\phi^l(T)<{a}^l$, which means that the input $\lambda^l/2L \leqslant {y}^{l-1}<\lambda^l-\lambda^l/2L$, the output of SNN is smaller than that of ANN (fires fewer spikes as expected).    
        \item[$\bullet$] \textbf{Case 4}: ${a}^l=\lambda^l,\phi^l(T)<{a}^l$, which means that  input ${y}^{l-1}\geqslant \lambda^l-\lambda^l/2L$, the output of ANN equals $\lambda^l$ but the output of SNN is smaller than $\lambda^l$ (fires fewer spikes as expected).
\end{itemize}
Note that here we use the scalars $a^l$ and $\phi^l(T)$ to represent the outputs of two corresponding neurons in ANN and SNN, respectively.

\subsection{Analyze the Distribution of Unevenness Error}
The discussion above of unevenness error ($\textbf{Error}^l$) in layer $l$ is based on the assumption that $\boldsymbol{a}^{l-1}=\boldsymbol{\phi}^{l-1}(T)$. However, in actual situations, $\boldsymbol{a}^{l-1}$ is not always equal to $\boldsymbol{\phi}^{l-1}(T)$, as unevenness error existed before the $l$-th layer has already caused the deviation between the respective output of ANNs and SNNs. Therefore, we attempt to give a more detailed discussion about unevenness error here.

\begin{definition}
\rm{Unevenness Error \uppercase\expandafter{\romannumeral1}}: $\boldsymbol{a}^{l-1}=\boldsymbol{\phi}^{l-1}(T)~\wedge~\boldsymbol{a}^{l} \neq \boldsymbol{\phi}^{l}(T)$, this condition is consistent with the original definition of unevenness error and represents the error completely generated by the $l$-th layer.
\end{definition}

\begin{definition}
\rm{Unevenness Error \uppercase\expandafter{\romannumeral2}}: $\boldsymbol{a}^{l} \neq \boldsymbol{\phi}^{l}(T)$ whether $\boldsymbol{a}^{l-1}$ equals $\boldsymbol{\phi}^{l-1}(T)$ or not. This condition reflects the cumulative effect of unevenness error.
\end{definition}

\begin{figure} [t]\centering    
\subfigure[Un. Error \uppercase\expandafter{\romannumeral1} (CIFAR-10)] {
 \label{fig0301}     
\includegraphics[width=0.47\columnwidth]{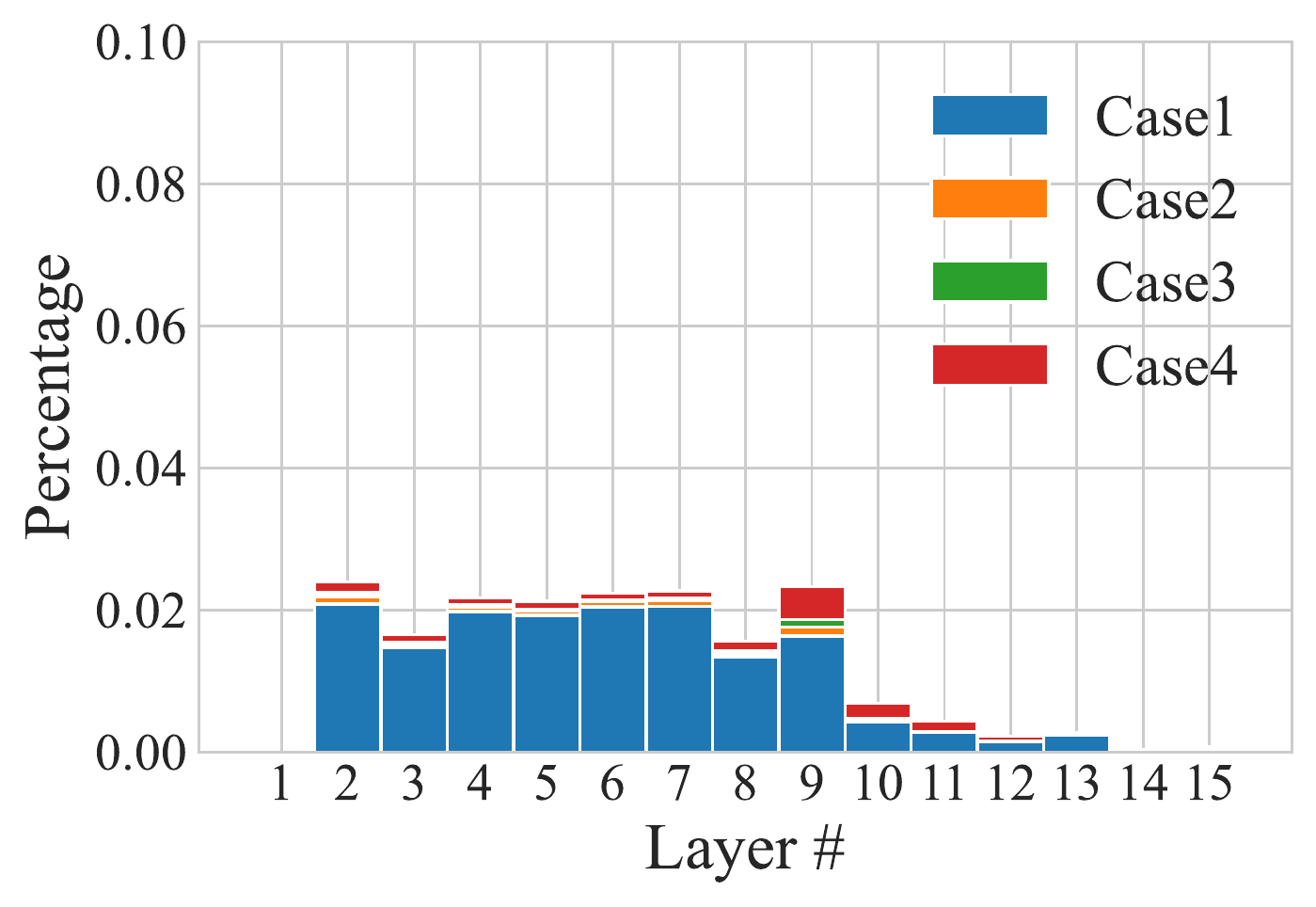}
}     
\subfigure[Un. Error \uppercase\expandafter{\romannumeral1} (CIFAR-100)] { 
\label{fig0302}     
\includegraphics[width=0.47\columnwidth]{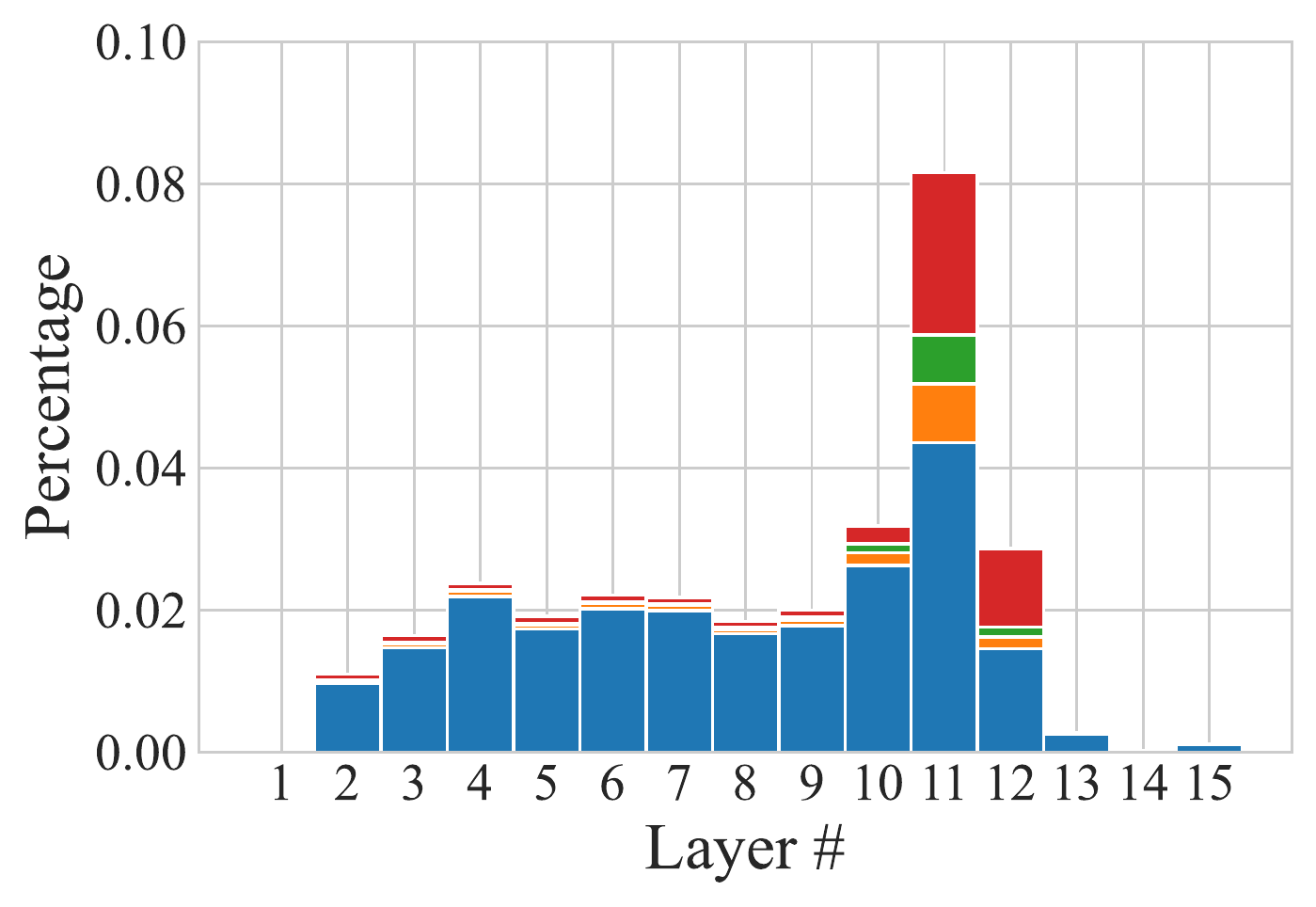}
}    
\subfigure[Un. Error \uppercase\expandafter{\romannumeral2} (CIFAR-10)] { 
\label{fig0303}     
\includegraphics[width=0.47\columnwidth]{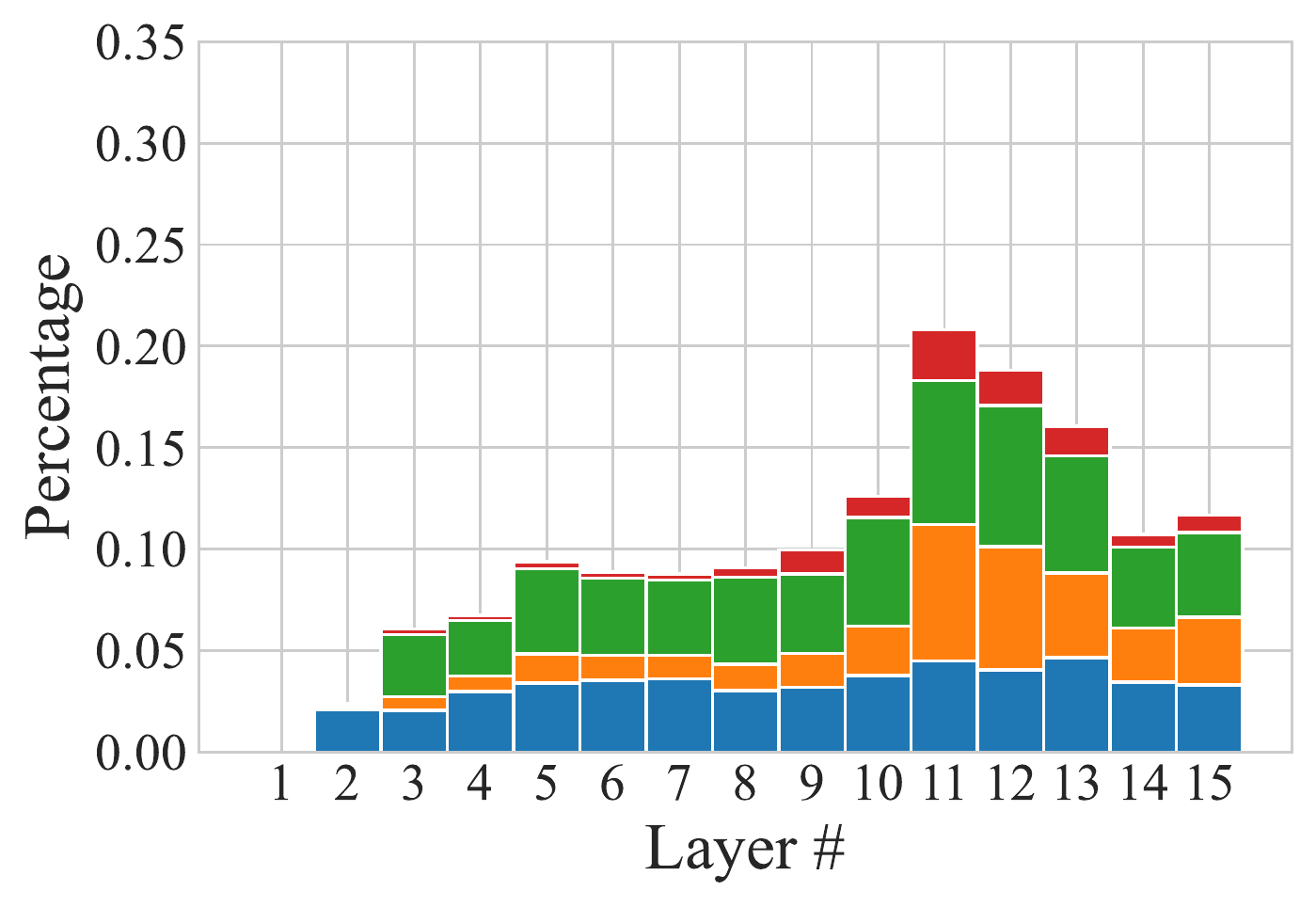}
}
\subfigure[Un. Error \uppercase\expandafter{\romannumeral2} (CIFAR-100)] { 
\label{fig0304}     
\includegraphics[width=0.47\columnwidth]{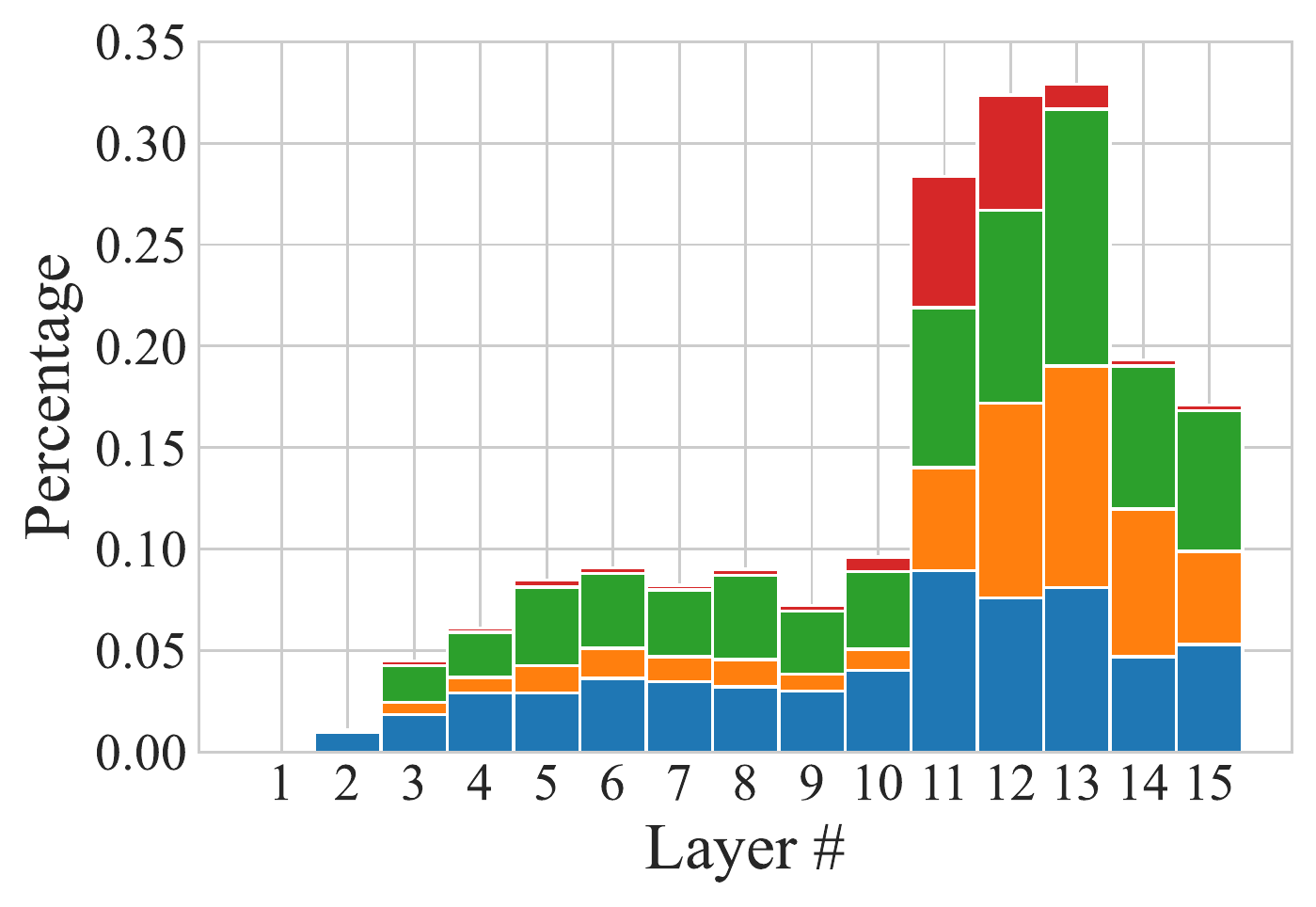}
}
\caption{The distribution of Unevenness Error \uppercase\expandafter{\romannumeral1}/\uppercase\expandafter{\romannumeral2} in each layer of VGG-16. (a)-(b):  Unevenness Error \uppercase\expandafter{\romannumeral1}, (c)-(d)  Unevenness Error \uppercase\expandafter{\romannumeral2}.}
\label{fig03}      
\end{figure}

For Unevenness Error \uppercase\expandafter{\romannumeral2} of the $l$-th layer, it can be considered as the joint effect of Unevenness Error \uppercase\expandafter{\romannumeral1} in the $l$-th layer (when $\boldsymbol{a}^{l-1}=\boldsymbol{\phi}^{l-1}(T)$) and Unevenness Error \uppercase\expandafter{\romannumeral2} in the $l-1$-th layer (when $\boldsymbol{a}^{l-1} \neq \boldsymbol{\phi}^{l-1}(T)$). Therefore, Unevenness Error \uppercase\expandafter{\romannumeral1} is the essential reason that affects the performance of SNNs, which desires to be eliminated.

We further analyze the distribution of Unevenness Error \uppercase\expandafter{\romannumeral1}/\uppercase\expandafter{\romannumeral2}  in four situations, and mine the main part of unevenness Error. Here, we train the source ANNs with QCFS activation function (Eq.~\eqref{equ08}) and VGG-16 structure on CIFAR-10/100 datasets and then convert them to SNNs. For unevenness Error \uppercase\expandafter{\romannumeral1} in layer $l$, we force the input $\boldsymbol{a}^{l-1}$ to ANN to be equal to the input $\boldsymbol{\phi}^{l-1}(T)$ to SNN, that is, $\boldsymbol{a}^{l-1}=\boldsymbol{\phi}^{l-1}(T)$, and recalculate the output $\boldsymbol{a}^{l}$ of ANN. Then we can compute the distribution of four kinds of Unevenness Error \uppercase\expandafter{\romannumeral1} by analyzing the values of $\boldsymbol{a}^{l}$ and $\boldsymbol{\phi}^{l}(T)$ of all neurons. For Unevenness Error \uppercase\expandafter{\romannumeral2} in layer $l$, we directly compute the distribution of four kinds of unevenness error by analyzing the values of $\boldsymbol{a}^{l}$ and $\boldsymbol{\phi}^{l}(T)$ of all neurons.
The results are illustrated in Fig.~\ref{fig03}, we have the following observation.\\
\textbf{Observation 1.} From Fig.~\ref{fig0303}-\ref{fig0304}, we note that all the four cases mentioned above have a remarkable effect on Unevenness Error \uppercase\expandafter{\romannumeral2}. However, for Unevenness Error \uppercase\expandafter{\romannumeral1} that determines Unevenness Error \uppercase\expandafter{\romannumeral2}, we can find that \textbf{Case 1} (${a}^l=0, \phi^l(T)> {a}^l$)  plays a more important role among four kinds of errors, which has a significant impact in almost every layer (Fig.~\ref{fig0301}-\ref{fig0302}).
Therefore, our main goal is to optimize this case to alleviate the situation of firing additional spikes.

\begin{figure} [t]\centering    
\includegraphics[width=0.8\columnwidth,trim=220 90 220 30,clip]{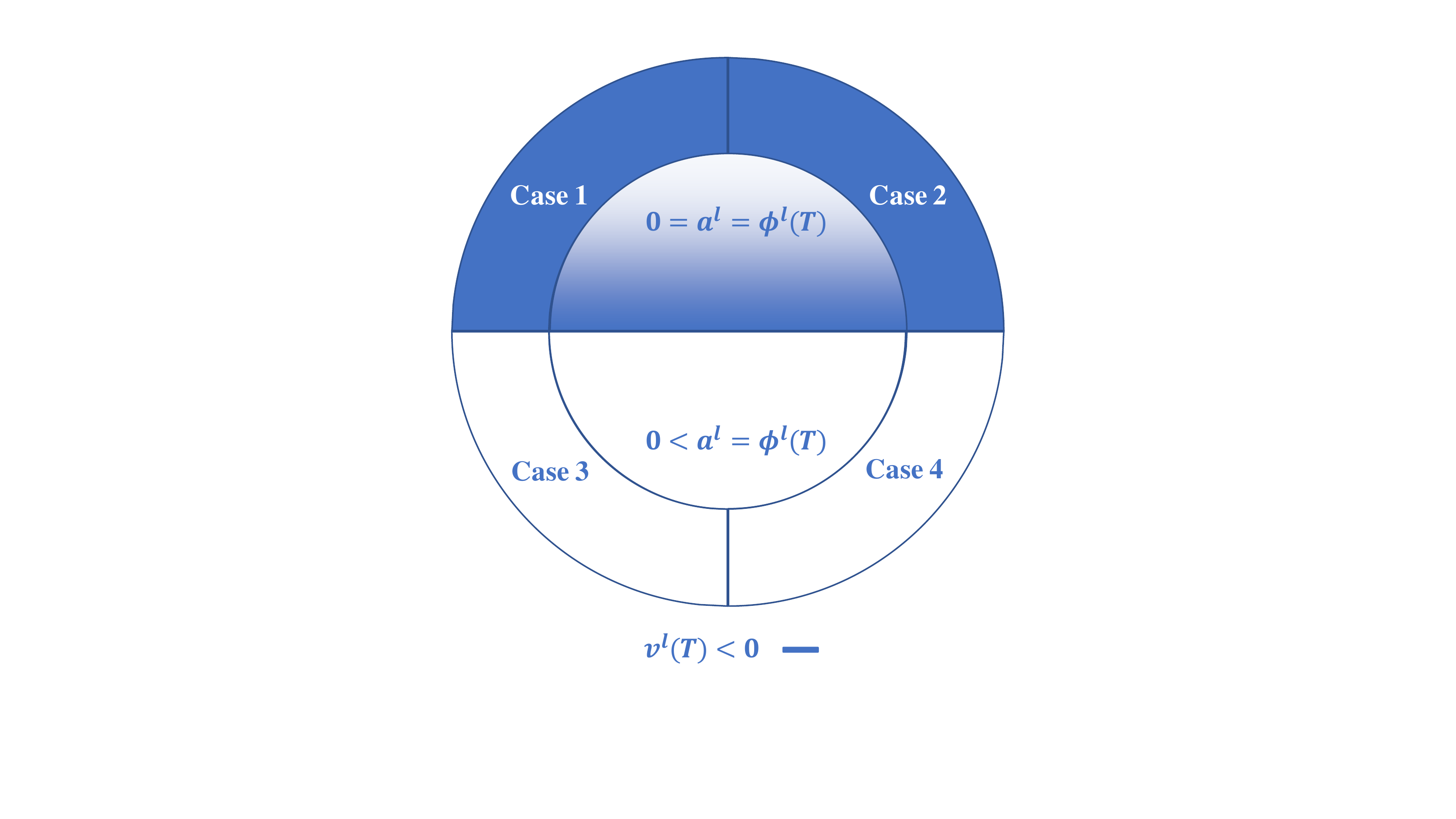}
\caption{Illustration of all cases of $\textbf{Error}^l$. The small concentric circle denotes that there is no error ($\textbf{Error}^l=0$), while the circular ring denotes the four cases of unevenness error. The semicircle (blue) represents all the cases covered by $\boldsymbol{v}^l(T)<0$.}
\label{residual_membrane_potential}     
\end{figure}

\subsection{Optimization Strategy Based on Residual Membrane Potential (SRP)}
Here we propose an optimization strategy to reduce unevenness error based on residual membrane potential. Our goal is to reduce Unevenness Error \uppercase\expandafter{\romannumeral1} of \textbf{Case 1} (${a}^l=0, \phi^l(T)> {a}^l$). However, in the practical application of SNNs, we cannot directly obtain the specific value of $\boldsymbol{a}^l$, which becomes an obstacle to further determining the category of unevenness Error \uppercase\expandafter{\romannumeral1}. Fortunately, we find that we can compare $\boldsymbol{a}^l$ with $\boldsymbol{\phi}^l(T)$ according to the value of residual membrane potential $\boldsymbol{v}^l(T)$, and we have the following theorem:

\begin{theorem}
Supposing that an ANN with QCFS activation (Eq.~\eqref{equ08}) is converted to an SNN with $L=T$ and $\lambda^l={\theta}^l$, and the ANN and SNN receive the same weighted input $\boldsymbol{y}^{l-1} = \boldsymbol{W}^{l} \boldsymbol{a}^{l-1}= \boldsymbol{W}^{l}\boldsymbol{\phi}^{l-1}(T)$,
 then we will have the following conclusions:\\
(\romannumeral1) If ${a}^l=0$, ${v}^l(T)<0$ is the sufficient condition of $\phi^l(T) \geqslant{a}^l$. In addition, ${v}^l(T)<0$ is also the necessary condition of $\phi^l(T)>{a}^l$.\\ 
(\romannumeral2) If ${a}^l>0$, ${v}^l(T)<0$ is the sufficient and necessary condition of $\phi^l(T)>{a}^l$.
\label{thm06}
\end{theorem}

The proof is provided in the appendix. Theorem 1 implies that if ${v}^l(T)<0$, we can infer that there is no error (${a}^l={\phi}^l(T)$) or there exist Unevenness Error \uppercase\expandafter{\romannumeral1} of \textbf{Case 1} and \textbf{Case 2} ($\phi^l(T)>{a}^l$), which is illustrated in the blue semicircle of Fig.~\ref{residual_membrane_potential}. However, we are unable to determine whether it belongs to \textbf{Case 1} (${a}^l=0, \phi^l(T)>{a}^l$) or \textbf{Case 2} ($0<{a}^l<\lambda^l,\phi^l(T)>{a}^l$). In fact, we have shown in Fig.~\ref{fig0301}-\ref{fig0302} that \textbf{Case 1} accounts for the largest percentage and \textbf{Case 2} can be approximately neglected. Thus we can conclude that when ${v}^l(T)< 0$, there exist mainly Unevenness Error \uppercase\expandafter{\romannumeral1} of \textbf{Case 1} or there is not any error.


\begin{algorithm}[t]
    \caption{The optimization strategy based on residual membrane potential (SRP)}
    \begin{algorithmic}[1]
        \REQUIRE Time-step to calculate residual membrane potential $\tau$; Time-step to test dataset $T$; Pretrained QCFS ANN model $f_\text{ANN}(\boldsymbol{W},\lambda)$; Dataset $D$.
        \ENSURE SNN model $f_\text{SNN}(\boldsymbol{W},\boldsymbol{v},\theta)$.
        \FOR{$l=1$ to $f_\text{ANN}.$layers}
            \STATE $f_\text{SNN}.{\theta}^l\gets f_\text{ANN}.\lambda^l$
            \STATE $f_\text{SNN}.\boldsymbol{v}^l(0)\gets\frac{1}{2}f_\text{SNN}.{\theta}^l$
            \STATE $f_\text{SNN}.\boldsymbol{W}^l\gets f_\text{ANN}.\boldsymbol{W}^l$
        \ENDFOR
        \FOR{length of Dataset $D$}
            \STATE Sample minibatch (data,label) from $D$
            \FOR{$t=1$ to $\tau$}
                \STATE $f_\text{SNN}($data$)$
            \ENDFOR
            \FOR{$l=1$ to $f_\text{SNN}.$layers}
                \STATE $f_\text{SNN}.{mask}^l \gets(f_\text{SNN}.\boldsymbol{v}^l(\tau)\geqslant 0)$
                \STATE $f_\text{SNN}.\boldsymbol{v}^l(\tau)\gets\frac{1}{2}f_\text{SNN}.{\theta}^l$
            \ENDFOR
            \FOR{$t=1$ to $T$}
                \FOR{$l=1$ to $f_\text{SNN}.$layers}
                    \STATE data $\gets f_\text{SNN}^l($data$)\cdot f_\text{SNN}.{mask}^l$
                \ENDFOR
            \ENDFOR
        \ENDFOR
        \RETURN $f_\text{SNN}(\boldsymbol{W},\boldsymbol{v},\theta)$
    \end{algorithmic}
    \label{algorithm01}
\end{algorithm}

Based on the above analysis, we can use the residual membrane potential to reduce unevenness error. If the residual membrane potential ${v}^l(T)$ is smaller than zero, we should set the corresponding output $\phi^l(T)$ to zero, so as to make $\phi^l(T)={a}^l=0$.
In practice, we first train an ANN with QCFS activation and then convert it to an SNN. In the inference stage, we use a two-stage strategy. For the first stage, we readout residual membrane potential  $\boldsymbol{v}^l(\tau)$ with $\tau$ denoting the actual time-step we choose.
If $\boldsymbol{v}^l(\tau)<0$ for a neuron, we set it to the dead neuron, which will not work in the next stage.
For the second stage, we run the SNN and report the result for a given time-step $T$.
 We name our optimization strategy as SRP in the following content. The specific algorithm flow about SRP is shown in Algorithm \ref{algorithm01}.

\begin{table*}[]
    \caption{Comparison between the proposed method and previous works on CIFAR-10 dataset}
    \renewcommand\arraystretch{1.15}
	\centering
        \begin{threeparttable}
	\begin{tabular}{cccccccccc}\hline
	
	Method & Arch & ANN & T=1 & T=2 & T=4 & T=8 & T=16 & T=32 & T=64\\ \hline
	
	RMP & VGG-16 & 93.63\% & - & - & - & - & - & 60.30\% & 90.35\% \\\hline
	RTS & VGG-16 & 95.72\% & - & - & - & - & - & 76.24\% & 90.64\% \\\hline
	TCL\tnote{*} & VGG-16 & 94.57\% & - & - & - & - & - & 93.64\% & 94.26\% \\\hline
	SNNC-AP & VGG-16 & 95.72\% & - & - & - & - & - & 93.71\% & 95.14\%  \\\hline
	OPI & VGG-16 & 94.57\% & - & - & - & 90.96\% & 93.38\% & 94.20\% & 94.45\%  \\\hline
	QCFS & VGG-16 & 95.52\% & 88.41\% & 91.18\% & 93.96\% & 94.95\% & 95.40\% & 95.54\% & 95.55\% \\\hline
	\textbf{SRP($\tau=4$)} & VGG-16 & 95.52\% & 93.80\% & 94.47\% & 95.32\% & 95.52\% & 95.44\% & 95.42\% & 95.40\% \\ \hline

	RTS & ResNet-18 & 95.46\% & - & - & - & - & - & 84.06\% & 92.48\% \\\hline
	SNNC-AP & ResNet-18 & 95.46\% & - & - & - & - & - & 94.78\% & 95.30\% \\\hline
	OPI & ResNet-18 & 96.04\% & - & - & - & 75.44\% & 90.43\% & 94.82\% & 95.92\%  \\\hline
	QCFS & ResNet-18 & 95.64\% & 88.84\% & 91.75\% & 93.83\% & 95.04\% & 95.56\% & 95.67\% & 95.63\% \\\hline
	\textbf{SRP($\tau=4$)} & ResNet-18 & 95.64\% & 94.59\% & 95.06\% & 95.25\% & 95.60\% & 95.55\% & 95.55\% & 95.58\% \\ \hline

	TSC & ResNet-20 & 91.47\% & - & - & - & - & - & - & 69.38\% \\\hline
	OPI & ResNet-20 & 92.74\% & - & - & - & 66.24\% & 87.22\% & 91.88\% & 92.57\%  \\\hline
	QCFS & ResNet-20 & 91.77\% & 62.43\% & 73.20\% & 83.75\% & 89.55\% & 91.62\% & 92.24\% & 92.35\% \\\hline
	\textbf{SRP($\tau=4$)} & ResNet-20 & 91.77\% & 86.37\% & 88.73\% & 90.51\% & 91.37\% & 91.64\% & 91.72\% & 91.80\% \\ \hline
	
	\end{tabular}
        \begin{tablenotes}
        \footnotesize
        \item[*] Self-implementation results.
        \end{tablenotes}
        \end{threeparttable}
	\label{table07}
\end{table*}

\begin{table*}[t]
    \caption{Comparison between the proposed method and previous works on CIFAR-100 dataset}
    \renewcommand\arraystretch{1.12}
	\centering
        \begin{threeparttable}
	\begin{tabular}{cccccccccc}\hline
	Method & Arch & ANN & T=1 & T=2 & T=4 & T=8 & T=16 & T=32 & T=64\\ \hline
	
	RTS                    & VGG-16 & 77.89\% & - & - & - & - & - & 7.64\% & 21.84\% \\\hline

	SNNC-AP                & VGG-16 & 77.89\% & - & - & - & - & - & 73.55\% & 76.64\% \\\hline
	TCL\tnote{*}                    & VGG-16 & 76.32\% & - & - & - & - & - & 52.30\% & 71.17\% \\\hline
	SNM                    & VGG-16 & 74.13\% & - & - & - & - & - & 71.80\% & 73.69\% \\\hline
	OPI                    & VGG-16 & 76.31\% & - & - & - & 60.49\% & 70.72\% & 74.82\% & 75.97\%  \\\hline
	QCFS                   & VGG-16 & 76.28\% & - & 63.79\% & 69.62\% & 73.96\% & 76.24\% & 77.01\% & 77.10\% \\\hline
	
	\textbf{SRP($\tau=4$)} & VGG-16 & 76.28\% & 71.52\% & 74.31\% & 75.42\% & 76.25\% & 76.42\% & 76.45\% & 76.37\% \\ \hline
	RMP                    & ResNet-20 & 68.72\% & - & - & - & - & - & 27.64\% & 46.91\% \\ \hline
	OPI                    & ResNet-20 & 70.43\% & - & - & - & 23.09\% & 52.34\% & 67.18\% & 69.96\%  \\\hline
	QCFS                   & ResNet-20 & 69.94\% & - & 19.96\% & 34.14\% & 55.37\% & 67.33\% & 69.82\% & 70.49\% \\\hline
	\textbf{SRP($\tau=4$)} & ResNet-20 & 69.94\% & 46.48\% & 53.96\% & 59.34\% & 62.94\% & 64.71\% & 65.50\% & 65.82\% \\ \hline
	\end{tabular}
        \begin{tablenotes}
        \footnotesize
        \item[*] Self-implementation results.
        \end{tablenotes}
        \end{threeparttable}
	\label{table02}
\end{table*}

\begin{table*}[t]
    \caption{Comparison between the proposed method and previous works on ImageNet dataset}
    \renewcommand\arraystretch{1.15}
	\centering
	\begin{tabular}{cccccccccc}\hline
	
	Method & Arch & ANN & T=1 & T=2 & T=4 & T=8 & T=16 & T=32 & T=64 \\ \hline
	SNNC-AP & VGG-16 & 75.36\% & - & - & - & - & - & 63.64\% & 70.69\% \\\hline
	SNM & VGG-16 & 73.18\% & - & - & - & - & - & 64.78\% & 71.50\% \\\hline
	OPI & VGG-16 & 74.85\% & - & - & - & 6.25\% & 36.02\% & 64.70\% & 72.47\% \\\hline
	QCFS & VGG-16 & 74.29\% & - & - & - & 19.12\% & 50.97\% & 68.47\% & 72.85\% \\\hline
	
	\textbf{SRP($\tau=14$)} & VGG-16 & 74.29\% & 50.37\% & 61.37\% & 66.47\% & 68.37\% & 69.13\% & 69.35\% & 69.43\% \\ \hline
	SNNC-AP & ResNet-34 & 75.66\% & - & - & - & - & - & 64.54\% & 71.12\% \\\hline
	QCFS & ResNet-34 & 74.32\% & - & - & - & 35.06\% & 59.35\% & 69.37\% & 72.35\% \\\hline
	\textbf{SRP($\tau=8$)} & ResNet-34 & 74.32\% & 57.78\% & 64.32\% & 66.71\% & 67.62\% & 68.02\% & 68.40\% & 68.61\% \\ \hline
	\end{tabular}
	\label{table03}
\end{table*}

\section{Experiments}
In this section, we evaluate the performance of our methods for image classification tasks on CIFAR-10~\cite{LeCun1998CIFAR10}, CIFAR-100~\cite{Krizhevsky2009CIFAR100} and ImageNet~\cite{Deng2009ImageNet} datasets under the network architecture of ResNet-18, ResNet-20, ResNet-34~\cite{he2016deep} and VGG-16~\cite{Simonyan2014VGG16}. 
 We compare our method with previous state-of-the-art ANN-SNN conversion methods, including RMP~\cite{han2020rmp}, 
 TSC~\cite{Han&Roy2020}, 
 RTS~\cite{deng2020optimal}, 
 OPI~\cite{Bu2022OPI}, SNNC-AP~\cite{li2021free}, TCL~\cite{ho2021tcl}, QCFS~\cite{bu2022optimal} and SNM~\cite{wang2022signed}.


\subsection{SRP Can Reduce Unevenness Error}
\begin{figure}[t] \centering    
\subfigure[VGG-16, CIFAR-10] {
 \label{fig1101}     
\includegraphics[width=0.9\columnwidth]{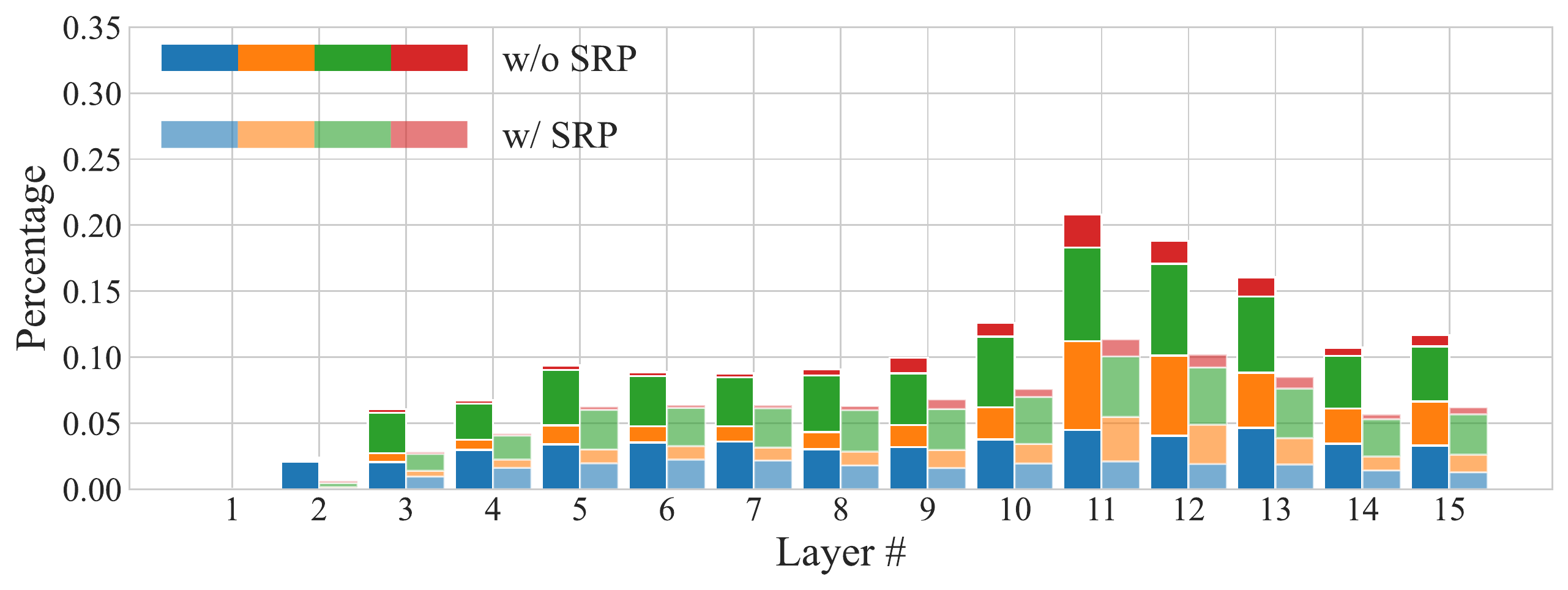}
}     
\subfigure[VGG-16, CIFAR-100] { 
\label{fig1103}     
\includegraphics[width=0.9\columnwidth]{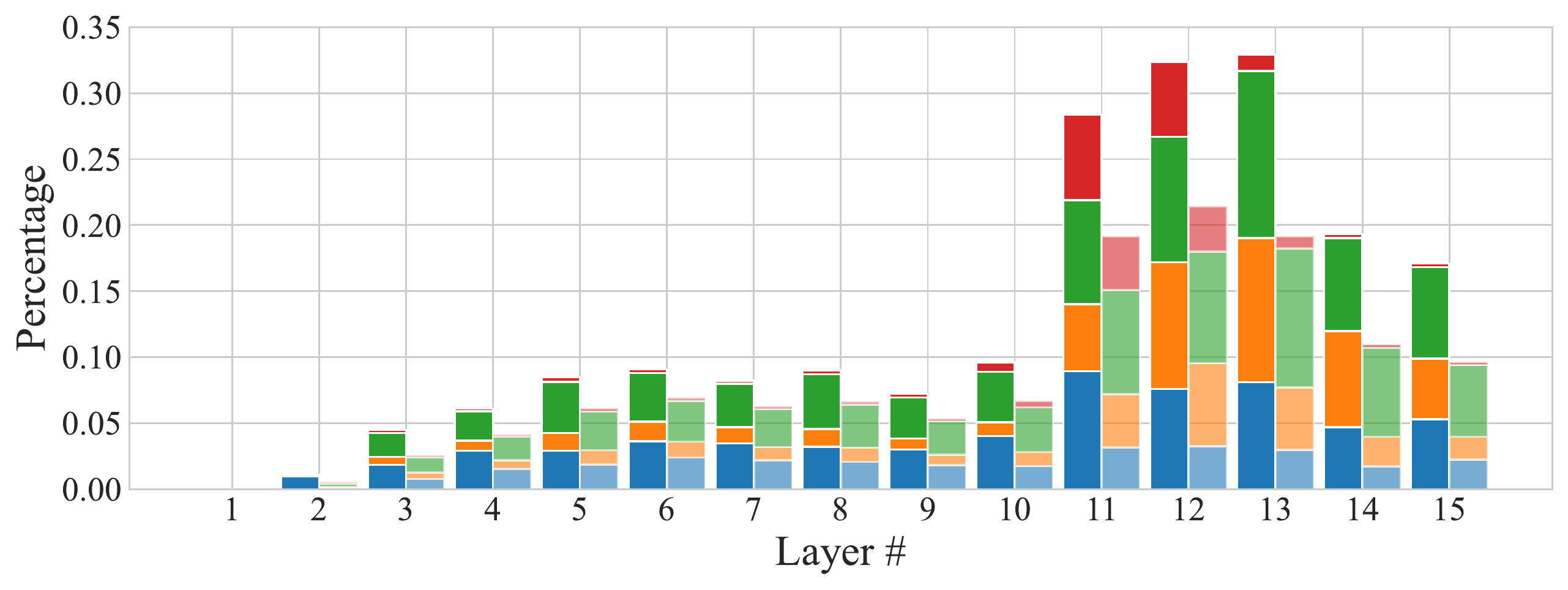}
}
\caption{Comparison of the distributions of Unevenness Error \uppercase\expandafter{\romannumeral2} before and after using SRP in each layer.}
\label{fig11}      
\end{figure}

We first test whether the proposed SRP can reduce unevenness error. We train VGG-16 on the CIFAR-10/100 datasets and then convert it to SNNs with/without SRP.
As Unevenness Error \uppercase\expandafter{\romannumeral2} can evaluate the overall performance of SNNs, we compare the distribution of Unevenness Error \uppercase\expandafter{\romannumeral2} before and after using SRP in each layer and show the results in Fig.~\ref{fig11}. One can find that our proposed method can reduce the percentage of unevenness Error \uppercase\expandafter{\romannumeral2} that existed in each layer significantly. For VGG-16 on CIFAR-10, SRP can reduce the percentage of unevenness Error \uppercase\expandafter{\romannumeral2} that existed in last-5 layers by 9.44\%, 8.60\%, 7.55\%, 5.06\%, and 5.47\%. For CIFAR-100, the corresponding decreased percentage  is 9.21\%, 10.96\%, 13.77\%, 8.34\%, and 7.48\%.

\begin{table*}[]
    \caption{Comparison about different $\tau$ of SRP on CIFAR-100 dataset}
    \renewcommand\arraystretch{1.15}
	\centering
	\begin{tabular}{cccccccccc}\hline
	Method & Arch & ANN & T=1 & T=2 & T=4 & T=8 & T=16 & T=32 & T=64\\ \hline
	Baseline & VGG-16 & 76.28\% & 57.50\% & 63.79\% & 69.62\% & 73.96\% & 76.24\% & 77.01\% & 77.10\% \\\hline
	\textbf{SRP($\tau=2$)} & VGG-16 & 76.28\% & 68.55\% & 71.24\% & 73.21\% & 74.16\% & 74.91\% & 75.13\% & 75.26\% \\ \hline
	\textbf{SRP($\tau=4$)} & VGG-16 & 76.28\% & 71.52\% & 74.31\% & 75.42\% & 76.25\% & 76.42\% & 76.45\% & 76.37\% \\ \hline
	\textbf{SRP($\tau=8$)} & VGG-16 & 76.28\% & 72.15\% & 75.20\% & 76.25\% & 76.66\% & 77.03\% & 77.14\% & 77.15\% \\ \hline
        \textbf{SRP($\tau=16$)} & VGG-16 & 76.28\% & 72.97\% & 75.66\% & 76.55\% & 77.03\% & 77.08\% & 77.08\% & 77.07\% \\ \hline  
    Baseline & ResNet-20 & 69.94\% & 13.19\% & 19.96\% & 34.14\% & 55.37\% & 67.33\% & 69.82\% & 70.49\% \\\hline
	\textbf{SRP($\tau=2$)} & ResNet-20 & 69.94\% & 30.99\% & 39.55\% & 47.72\% & 55.13\% & 59.41\% & 60.76\% & 61.80\% \\ \hline
	\textbf{SRP($\tau=4$)} & ResNet-20 & 69.94\% & 46.48\% & 53.96\% & 59.34\% & 62.94\% & 64.71\% & 65.50\% & 65.82\% \\ \hline
	\textbf{SRP($\tau=8$)} & ResNet-20 & 69.94\% & 55.29\% & 63.07\% & 66.44\% & 68.22\% & 68.93\% & 68.95\% & 68.96\% \\ \hline
	\textbf{SRP($\tau=16$)} & ResNet-20 & 69.94\% & 58.32\% & 65.16\% & 68.07\% & 69.50\% & 69.27\% & 69.47\% & 69.49\% \\ \hline
	\end{tabular}
	\label{table06}
\end{table*}

\subsection{Comparison with the State-of-the-art}
We compare our method with the state-of-the-art ANN-SNN conversion methods to validate the effectiveness of our method. For fair comparison, we use the performance of SRP on $T=t$ to compare with the performance of other state-of-the-arts on $T=t+\tau$.

Tab.~\ref{table07} shows the performance of our proposed method on the CIFAR-10 dataset. As CIFAR-10 is a relatively easy dataset, the advantages of SRP are not as remarkable as those on CIFAR-100 and ImageNet dataset. For VGG-16 architecture, we achieve 95.32\% with 4 time-steps ($\tau=4$), which is 4.36\% higher than OPI and 0.37\% higher than QCFS on $T=8$. The accuracy of our method ($\tau=4$) on $T=4$ has outperformed the performance of other previous works on $T=64$, which include RMP, TCL, RTS, SNM and SNNC-AP. For ResNet-18, we obtain 95.25\% after using 4 time-steps ($\tau=4$), which is 19.81\% higher than OPI and 0.21\% higher than QCFS on $T=8$ respectively. For ResNet-20, the performance of SRP ($\tau=4$) on $T=4$ can outperform OPI and QCFS on $T=8$ with 24.27\% and 0.96\%.

Tab.~\ref{table02} reports the results on the CIFAR-100 dataset. For low latency inference ($T<16$), our model outperforms all the other methods with the same time-step setting. For VGG-16, we achieve 75.42\% top-1 accuracy with 4 time-steps ($\tau=4$), whereas the methods of OPI and QCFS
reach 60.49\% and 73.96\% accuracy at the end of 8 time-steps. Moreover, we achieve an accuracy of 71.52\% using only 1 time-step. For ResNet-20, the proposed method achieve 59.34\% top-1 accuracy with 4 time-steps ($\tau=4$), which is 36.25\% higher than OPI and 3.97\% higher than QCFS, respectively.

We further evaluate whether our method can be generalized to the large-scale dataset. Here we test the performance of our method on the ImageNet dataset and report the results in Tab.~\ref{table03}. One can find that our method shows its superiority more obviously in the large-scale dataset. For VGG-16, SRP achieves the accuracy of 61.37\% with 2 time-steps ($\tau=14$), whereas the methods of OPI and QCFS reach 36.02\% and 50.97\% accuracy at the end of 16 time-steps. For ResNet-34, we obtain 67.62\% accuracy with 8 time-steps ($\tau=8$), which outperforms QCFS (59.35\% when $T=16$) by 8.27\% accuracy. All these results demonstrate that our method outperforms the previous conversion methods.

\subsection{The Effect of Parameter $\tau$}
Here we test the effects of parameter $\tau$  in our SRP. There is a trade-off between performance and latency. Larger $\tau$ can make a more accurate estimation of the residual membrane potential, thereby increasing the performance of SNNs. However, these advantages come with the price of long latency. If $\tau$ is set much smaller than the quantization step $L$ in the QCFS activation function of ANN, the precondition $\boldsymbol{y}^{l-1} = \boldsymbol{W}^{l} \boldsymbol{a}^{l-1}= \boldsymbol{W}^{l}\boldsymbol{\phi}^{l-1}(T)$ in Theorem 1 will be violated, which will decrease the performance of SRP.

\begin{figure}[t] \centering    
\subfigure[VGG-16 on CIFAR-100] { 
\label{fig0601}     
\includegraphics[width=0.45\columnwidth,trim=220 100 220 100,clip]{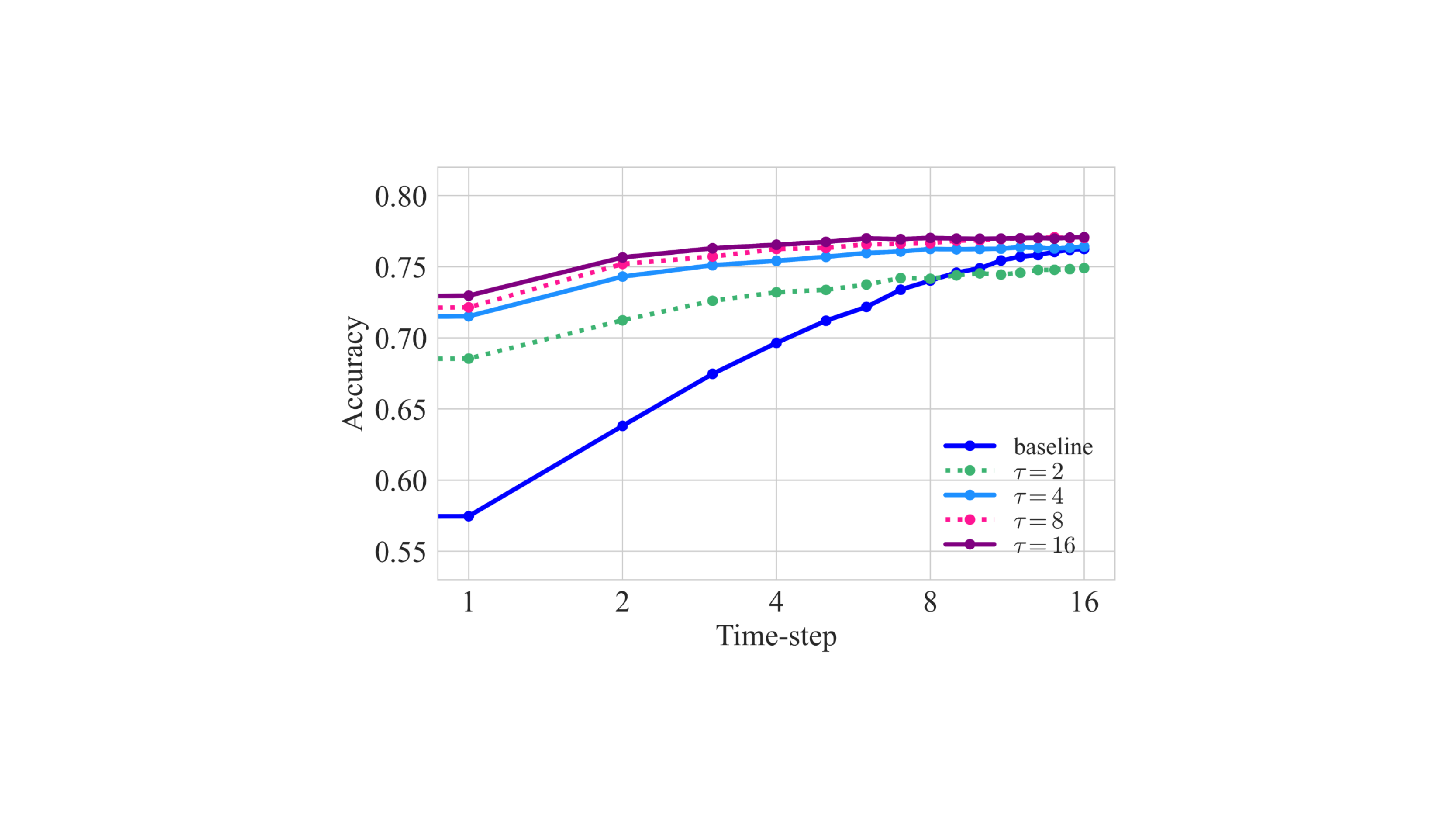} 
}   
\subfigure[ResNet-20 on CIFAR-100] { 
\label{fig0602} 
\includegraphics[width=0.45\columnwidth,trim=220 100 220 100,clip]{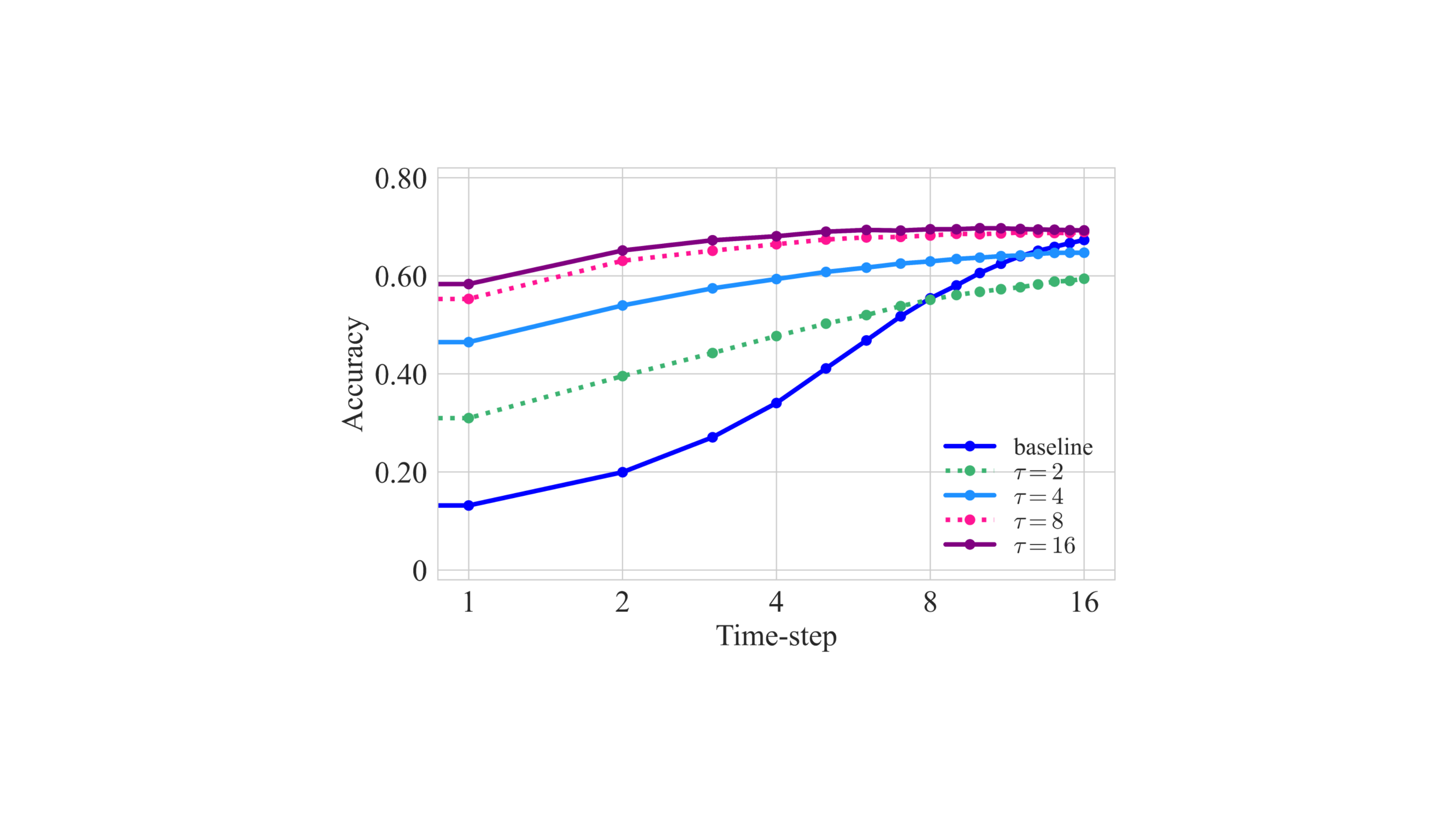}
} 
\subfigure[VGG-16 on ImageNet] { 
\label{fig0603}     
\includegraphics[width=0.45\columnwidth,trim=220 100 220 100,clip]{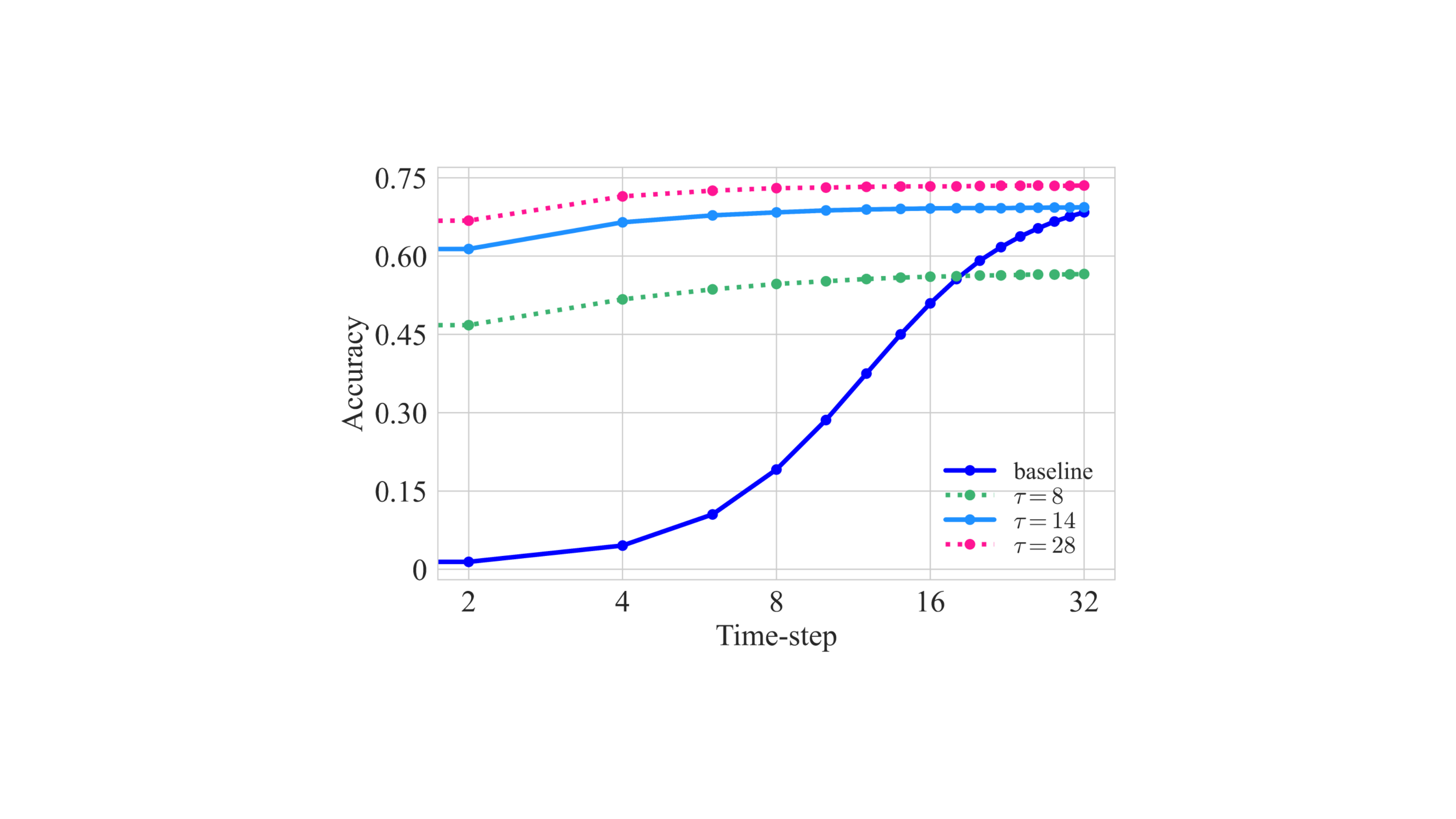} 
}   
\subfigure[ResNet-34 on ImageNet] { 
\label{fig0604} 
\includegraphics[width=0.45\columnwidth,trim=220 100 220 100,clip]{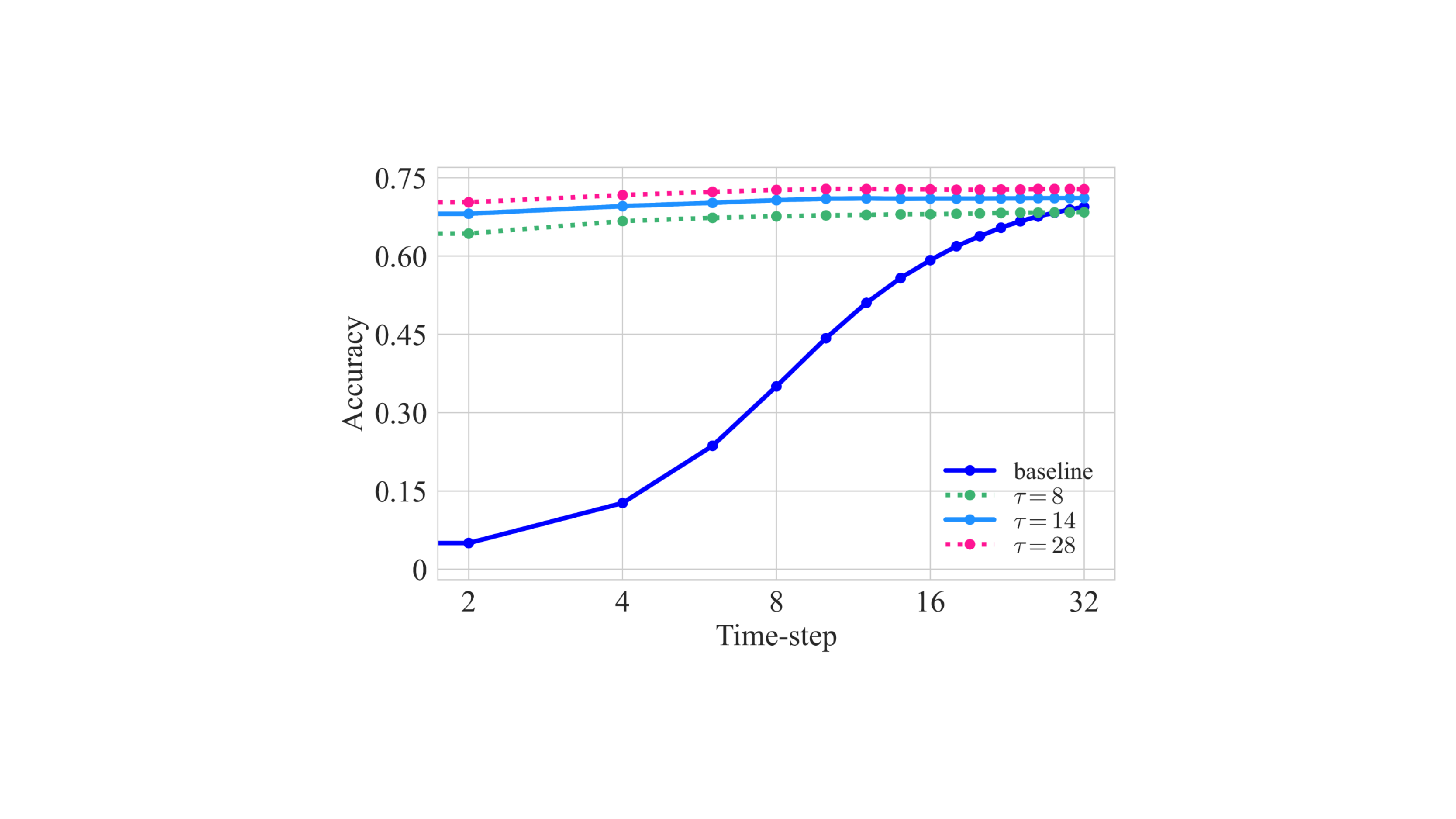}
}  
\caption{The effect of parameter $\tau$}     
\label{fig06}     
\end{figure}

For a more detailed analysis, we train a source ANN on CIFAR-100/ImageNet dataset and then convert it to SNNs with different $\tau$. The performance of converted SNNs is shown in Fig.~\ref{fig06} and Tab.~\ref{table06}. For CIFAR-100, the quantization step $L$  is set to 4 on VGG-16 and 8 on ResNet 20. For ImageNet, the quantization step $L$ is set to 16 on VGG-16 and 8 on ResNet-34. One can find that as $\tau$ increases, the accuracy of SNNs generally improves. When $\tau$ is nearly equal to $L$, the performance of SNNs stabilizes gradually, which implies that a too large $\tau$ will increase latency but cannot improve the accuracy remarkably. Besides, the results demonstrate that despite a significant change in $\tau$, our SRP method still maintains good generalization capabilities.

In addition, for the tasks with different allowable time-steps, we can set different $\tau$ to improve the performance of SNNs. Specifically, we consider using relatively small $\tau$ under low time latency and use larger $\tau$ for a longer time-step.

\subsection{Apply SRP to Other ANN-SNN Models}
\begin{figure} \centering
\subfigure[TCL on CIFAR-100] { 
\label{fig1001}     
\includegraphics[width=0.47\columnwidth,trim=200 100 200 100,clip]{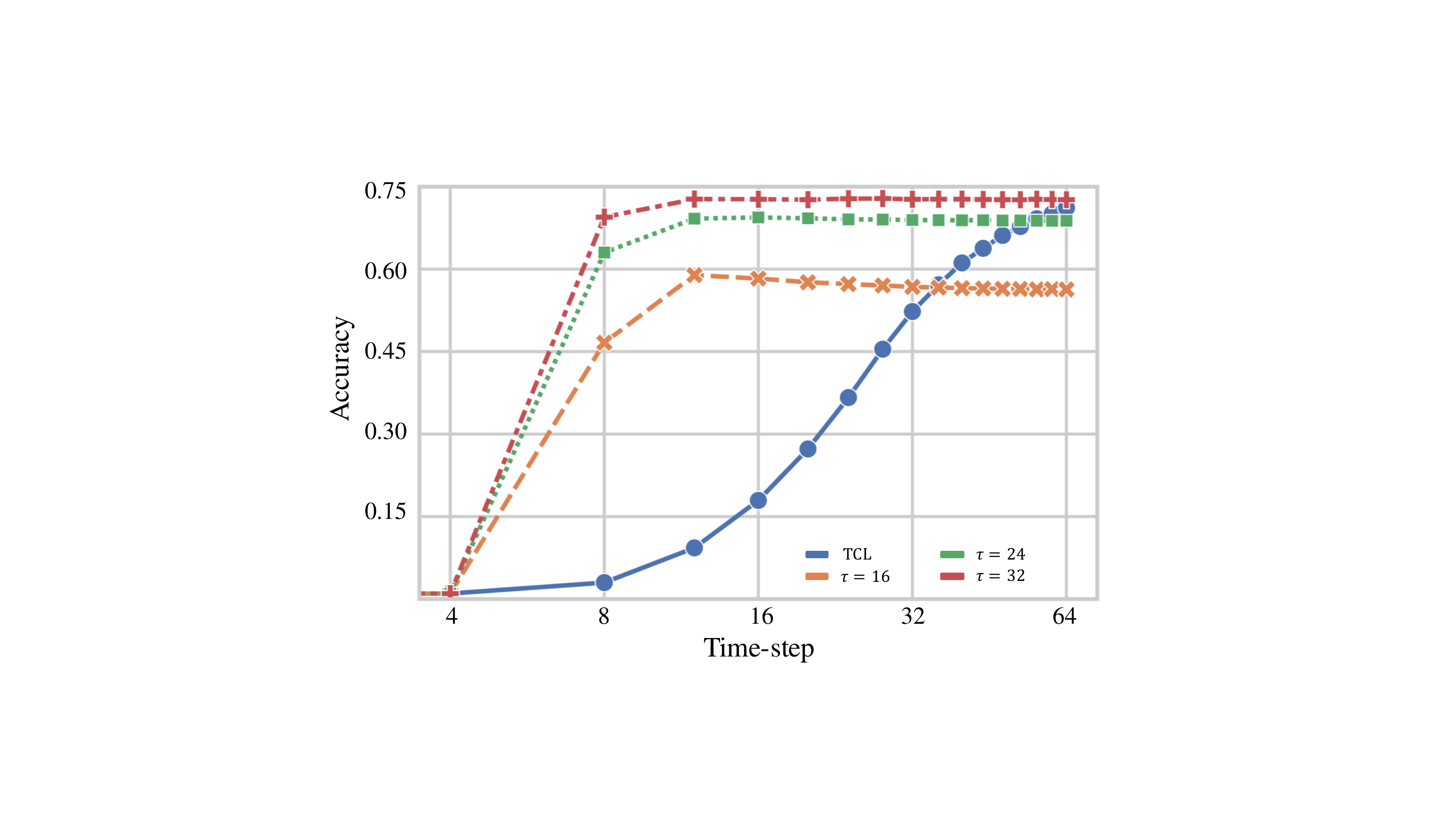}
}    
\subfigure[OPI on CIFAR-100] { 
\label{fig1005} 
\includegraphics[width=0.47\columnwidth,trim=200 100 200 100,clip]{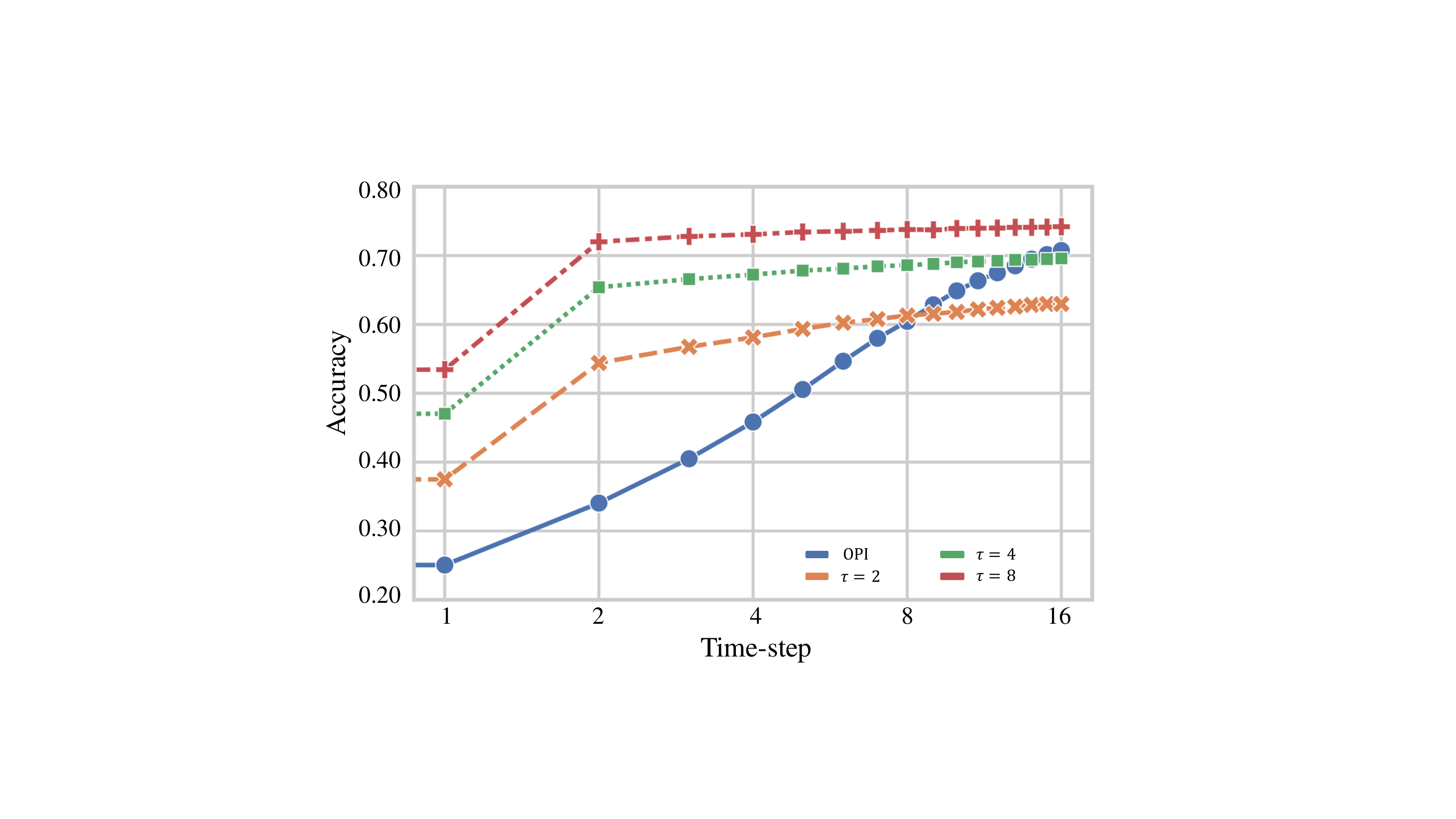}
}   
\caption{Effect of SRP on other models}
\label{fig10}     
\end{figure}

In this section, we experimentally prove that SRP can be generalized to other conversion methods. To verify this, we consider TCL~\cite{ho2021tcl} and OPI~\cite{Bu2022OPI} as the basic conversion method and then add SRP to those methods. 
As shown in Fig.~\ref{fig10}, we can find that the performance of both TCL and OPI increases after using SRP. Specifically, for TCL model, we achieve 63.03\% with 8 time-steps ($\tau=24$), which outperforms baseline (52.30\% when $T=32$) with 10.73\% accuracy. For OPI model, we obtain 67.24\% with 4 time-steps ($\tau=4$), which is 6.79\% higher than baseline (60.45\% when $T=8$). These results show that SRP can apply to other conversion models and remarkably improve classification accuracy.

\subsection{Training Procedure and Hyperparameter Configuration}

The training procedure of ANNs is consistent with~\cite{bu2022optimal}. We replace ReLU activation layers with QCFS and select average-pooling as the pooling layers. We set $\lambda^l$ in each activation layer as trainable threshold and $\boldsymbol{v}^l(0)=\frac{1}{2}\theta^l$ in IF Neuron. For CIFAR-10, we set $L=4$ on all network architectures. For CIFAR-100, we set $L=4$ for VGG-16 and $L=8$ for ResNet-20. For ImageNet, we set $L=16$ for VGG-16 and $L=8$ for ResNet-34.

We use Stochastic Gradient Descent \cite{bottou2012stochastic} as our training  optimizer and cosine decay scheduler \cite{loshchilov2016sgdr} to adjust the learning rate. We set the momentum parameter as $0.9$. The initial learning rate is set as $0.1$ (CIFAR-10/ImageNet) or $0.02$ (CIFAR-100) and weight decay is set as $5\times10^{-4}$ (CIFAR-10/100) or $1\times10^{-4}$ (ImageNet). In addition, we also use common data normalization and data enhancement techniques \cite{devries2017improved,Ekin2019,li2021free} for all datasets. During the procedure of adopting SRP, for CIFAR-10/100, we recommend $\tau=4$ for all network architecture. For ImageNet, we recommend $\tau=14$ for VGG-16 and $\tau=8$ for ResNet-34.

\section{Conclusion}
In this paper, we systematically analyze the cases and distribution of unevenness error and propose an optimization strategy based on residual membrane potential to reduce unevenness error. By introducing the temporal information of SNNs, our method achieves state-of-the-art accuracy on CIFAR-10, CIFAR-100, and ImageNet datasets, with fewer time-steps. Our method can alleviate the performance gap between ANNS and SNN, and promote the practical application of SNNs on 
neuromorphic chips.

\appendix
\section{Appendix}

\subsection{Eliminating Clipping and Quantization Errors}
In the main text, we claim that the QCFS activation function (Eq.~(7)) better approximates the activation function of SNNs and could eliminate clipping and quantization errors. Here we give the detailed proof.\\
\textbf{Eliminating Clipping error.} Since the  QCFS activation function of Eq.~(7) includes the clip operation, we can know that the value range of $\boldsymbol{a}^l$ is $[0,\lambda^l]$. Therefore, the interval $[\lambda^l,{a}_{max}^l]$ has been clipped out and the clipping error is completely eliminated.\\
\textbf{Eliminating Quantization error.} According to \cite{bu2022optimal}, if we take no account of unevenness error, we can consider that the timing of receive spikes is even, then we will have:
\begin{align}
    \boldsymbol{\phi}^l(T) &= {\rm clip} \left(\frac{\theta^l}{T}\left\lfloor\frac{\boldsymbol{y}^{l-1} T+\boldsymbol{v}^l(0)}{\theta^l}\right\rfloor,0,\theta^l \right).
\end{align}
where we use $\boldsymbol{y}^{l-1} = \boldsymbol{W}^{l} \boldsymbol{a}^{l-1}= \boldsymbol{W}^{l}\boldsymbol{\phi}^{l-1}(T)$
to substitute the weighted input to layer $l$ for both ANN and SNN. When we use QCFS function, we generally set $\lambda^l=\theta^l,\boldsymbol{v}^l(0)=\theta^l/2$. As we do not consider unevenness error here and clipping error has already been eliminated, it is reasonable to believe that the deviation between $\boldsymbol{a}^l$ and $\boldsymbol{\phi}^l(T)$ here is mainly caused by quantization error. Therefore, we use $\boldsymbol{a}^l - \boldsymbol{\phi}^l(T)$ to represent the degree of quantization error. When $\boldsymbol{y}^{l-1}\in[0,\theta^l]$, $\forall T,L$ we can derive the following conclusion:
\begin{align}
    &\ \mathbb{E}\left(\boldsymbol{a}^l-\boldsymbol{\phi}^l(T)\right) \nonumber \\ 
    = &\ \mathbb{E}\left(\frac{\lambda^l}{L}\left\lfloor\frac{\boldsymbol{y}^{l-1}L}{\lambda^l}+\frac{1}{2}\right\rfloor - \frac{\theta^l}{T}\left\lfloor\frac{\boldsymbol{y}^{l-1} T+\boldsymbol{v}^l(0)}{\theta^l}\right\rfloor \right) \nonumber \\
    = &\ 0. \label{eq09}
\end{align}
The conclusion in Eq.~\eqref{eq09} has already been proved by \cite{bu2022optimal}. From the perspective of mathematical expectation, we have proved that the QCFS function can eliminate quantization error effectively.

\subsection{Proof of Theorem}

\textbf{Theorem 1.} {\it Supposing that an ANN with QCFS activation (Eq. (7)) is converted to an SNN with $L=T$ and $\lambda^l={\theta}^l$, and the ANN and SNN receive the same weighted input $\boldsymbol{y}^{l-1} = \boldsymbol{W}^{l} \boldsymbol{a}^{l-1}= \boldsymbol{W}^{l}\boldsymbol{\phi}^{l-1}(T)$,
 then we will have the following conclusions:\\
(\romannumeral1) If ${a}^l=0$, ${v}^l(T)<0$ is the sufficient condition of $\phi^l(T) \geqslant{a}^l$. In addition, ${v}^l(T)<0$ is also the necessary condition of $\phi^l(T)>{a}^l$.\\ 
(\romannumeral2) If ${a}^l>0$, ${v}^l(T)<0$ is the sufficient and necessary condition of $\phi^l(T)>{a}^l$.}

\begin{proof}

\textbf{The proof of necessity in (\romannumeral1).} By combining Eq.\eqref{equ07}, $\boldsymbol{v}^l(0)=\frac{{\theta}^l}{2}$ and $\boldsymbol{y}^{l-1}=\boldsymbol{W}^l\boldsymbol{\phi}^{l-1}(T)$, we will have
\begin{align}
    \phi^l(T) &= {y}^{l-1} - \frac{1}{T}\left({v}^l(T)-\frac{{\theta}^l}{2}\right). \label{eq08}
\end{align}
On the one hand, when ${a}^l=0$, according to the preconditions $L=T,\lambda^l={\theta}^l$ mentioned in Theorem 1 and the definition of QCFS activation function in Eq.\eqref{equ08}, we can know that ${y}^{l-1}<\frac{{\theta}^l}{2T}$. On the other hand, as $\phi^l(T)>{a}^l=0$, then we will have 
\begin{align}
    &\phi^l(T)= {y}^{l-1} - \frac{1}{T}\left({v}^l(T)-\frac{{\theta}^l}{2}\right) \geqslant \frac{{\theta}^l}{T},\  {y}^{l-1}<\frac{{\theta}^l}{2T}. \label{eq07}
\end{align}
According to Eq.~\eqref{eq07}, $\frac{1}{T}({v}^l(T)-\frac{{\theta}^l}{2})<-\frac{{\theta}^l}{2T}$, then we can derive ${v}^l(T)<0$.

\textbf{The proof of sufficiency in (\romannumeral1).} As ${\phi}^l(T)$ belongs to a finite discrete set $S_T = \{ \frac{{\theta}^l i}{T} | i\in [0,T] \wedge  i\in \mathbb{N} \}$, ${\phi}^l(T)\geqslant 0$ at any time. If ${a}^l=0$, $\phi^l(T) \geqslant{a}^l$.

\textbf{The proof of necessity in (\romannumeral2).} For ${a}^l>0$, we can set ${a}^l=\frac{k{\theta}^l}{T},k\in\mathbb{N}\wedge 0<k<T$, then we can have $y^{l-1}\in \left[\frac{{\theta}^l}{T}(k-\frac{1}{2}),\frac{{\theta}^l}{T}(k+\frac{1}{2})\right)$. When ${a}^l<\phi^l(T)$, from Eq.\eqref{eq08} we can have 
\begin{align}
    &\phi^l(T)= {y}^{l-1} - \frac{1}{T}\left({v}^l(T)-\frac{{\theta}^l}{2}\right) \geqslant \frac{(k+1){\theta}^l}{T}, \nonumber \\
    &y^{l-1}\in \left[\frac{{\theta}^l}{T}(k-\frac{1}{2}),\frac{{\theta}^l}{T}(k+\frac{1}{2})\right). \label{eq06}
\end{align}
From Eq.~\eqref{eq06}, it is obvious that $\frac{1}{T}({v}^l(T)-\frac{{\theta}^l}{2})<-\frac{{\theta}^l}{2T}$, which means that ${v}^l(T)<0$.

\textbf{The proof of sufficiency in (\romannumeral2).} When $a^l=\theta^l$, $y^{l-1}\geqslant\frac{{\theta}^l}{T}(T-\frac{1}{2})$. By combining ${v}^l(T)<0$ with $y^{l-1}\geqslant\frac{{\theta}^l}{T}(T-\frac{1}{2})$, from Eq.\eqref{eq08} we can have $\phi^l(T)>\theta^l$, which is impossible. Therefore, $a^l$ must be smaller than $\theta^l$.

When $0<a^l<\theta^l$, By combining ${v}^l(T)<0$ with ${a}^l=\frac{k{\theta}^l}{T}, y^{l-1}\in \left[\frac{{\theta}^l}{T}(k-\frac{1}{2}),\frac{{\theta}^l}{T}(k+\frac{1}{2})\right)$, we will have:
\begin{align}
    \phi^l(T) &= {y}^{l-1} - \frac{1}{T}\left({v}^l(T)-\frac{{\theta}^l}{2}\right) \nonumber \\
    &> {y}^{l-1} + \frac{\theta^l}{2T} \geqslant \frac{k{\theta}^l}{T} = {a}^l.
\end{align}
\end{proof}

\section{Acknowledgments}
This work was supported by the National Natural Science Foundation of China under Grant No. 62176003 and No. 62088102.

\bibliography{aaai23}

\end{document}